\pdfoutput=1
\documentclass{article}

\usepackage[preprint, nonatbib]{neurips_2026}
\usepackage{biblatex}
\usepackage{xr}
\usepackage{enumitem}
\usepackage{listings}

\usepackage[utf8]{inputenc}
\usepackage[T1]{fontenc}
\usepackage{hyperref}
\usepackage{url}
\usepackage{booktabs}
\usepackage{amsfonts}
\usepackage{nicefrac}
\usepackage{microtype}
\usepackage{xcolor}
\usepackage{graphicx}
\usepackage{amssymb}
\usepackage{placeins}
\usepackage{tikz}
\usepackage{bbm}
\usetikzlibrary{arrows.meta,calc}

\usepackage[normalem]{ulem}

\title{A\textsuperscript{4}-MR: Abnormality Aware Anatomical Alignment for MR}
\hypersetup{pdftitle={A4-MR: Abnormality Aware Anatomical Alignment for MR}}

\usepackage{authblk}

\author[1,*]{Noel Dsouza}
\author[1]{Zhijian Yang\textsuperscript{\textdagger}}
\author[1]{Nishank Singla}
\author[1]{Pranay Doshi\textsuperscript{\textdagger}}
\author[2]{Krisztian Koos}
\author[2]{Fei Tan}
\author[2]{Bruno Nunes}
\author[2]{Laszlo Rusko}
\author[2]{Bernadett Kolozsvari}
\author[2]{Eszter Sabo}
\author[1]{Emanuele Valeriano}
\author[1]{Hugh Pemberton}
\author[1,3]{Vikash Gupta}
\author[1]{Seyed Iman Zare Estakhraji}
\author[1]{Marc Lebel}
\author[1]{Farzin Haddadpour\textsuperscript{\textdagger}}
\author[1]{Taha Kass-Hout}
\author[1,*]{Erhan Bas}
\affil[1]{GE Healthcare, Seattle, USA}
\affil[2]{GE Healthcare, Budapest, Hungary}
\affil[3]{Amazon, USA}
\affil[*]{Corresponding author:
\linebreak Noel Dsouza; Erhan Bas,
\texttt{\{noel.dsouza, erhan.bas\}@gehealthcare.com}}

\begin{document}

\maketitle

\begingroup\renewcommand{\thefootnote}{\textdagger}\footnotetext{Work performed while at GE Healthcare.}\endgroup

\begin{abstract}
With the advent of large scale foundation models, significant progress has been made on their application to Magnetic Resonance Imaging.
Recent vision language MR models generalize well across a wide variety of clinical and research based MR tasks and perform on par with task-specific models.
However, we find anatomy specific localization and pathological analysis still remains lacking, especially in any real-world scenarios. To this end, we propose A\textsuperscript{4}-MR,
a 3D vision-language foundation model trained on around 50k MR studies (491k MR series), about half with corresponding radiology reports.
We aim to solve the following tasks, 1) localize anatomies through large scale pretraining leveraging noisy labels, 2) align image and text representations at the anatomy level,
3) make the anatomy-level alignment abnormality-aware. We demonstrate that our model surpasses existing foundation models on a diverse
set of tasks~{(e.g. +1.9 AUROC on CirrMRI600+ abnormality detection over the strongest baseline)}. Pooling the backbone tokens inside the predicted anatomy box separates abnormal from healthy organs across tasks (liver cirrhosis, Alzheimer's disease, prostate cancer), and the abnormality-aware targets further improve this on liver (+2.8 AUROC at 10\% labels). With a frozen encoder, it also outperforms end-to-end task-specific models on knee segmentation ({+1.2} Dice against MedNeXt) and is on par with them on abdominal organ segmentation (within 0.4 Dice of the best model on AMOS). Without any prompt, it localizes 3D anatomies as well as MedSAM2 given a ground-truth box.

\end{abstract}

\section{Introduction}
Magnetic resonance imaging (MRI) has transformed medical imaging by enabling safe, non-invasive visualization of soft tissue across a wide range of clinical tasks, including disease classification, organ-at-risk segmentation, and tumor identification and monitoring. Despite these advantages, automated MR analysis remains challenging. MR scans vary substantially across scanner hardware, acquisition protocols, spatial resolutions, and anatomical regions, while clinically relevant pathologies are often subtle and difficult to distinguish. These challenges are further compounded by the limited availability of large, diverse datasets with high-quality annotations.

In recent years, foundation models have emerged as a promising strategy for addressing these limitations. By leveraging large-scale unlabeled data, these models learn general-purpose representations that can be transferred to downstream tasks. For example, BERT and GPT-2~\cite{Devlin2019BERTPO,radford2019language} transformed natural language processing, while ViT~\cite{dosovitskiy2021an} played a similar role in computer vision. Self-supervised methods such as DINOv2~\cite{oquab2024dinov} and MAE~\cite{He2021MaskedAA} have further enabled large visual models to learn transferable image representations from unlabeled data at scale. In parallel, localization and detection methods such as YOLO and DETR~\cite{Redmon2015YouOL,Carion2020EndtoEndOD} have demonstrated strong progress in object-level visual understanding, with DETR showing the effectiveness of transformer architectures for detection.

These advances have increasingly been adapted to medical imaging, where foundation models have pushed performance in modalities such as CT~\cite{Blankemeier2024MerlinAC,Pai2025VisionFM}, X-ray~\cite{ma2025chestai,Singla2025MultiAX, Zhang2020ContrastiveLO}, and MR~\cite{Li2024TowardsAH,Yang2025DecipherMRAV,Sellergren2025MedGemmaTR}. Prior work has explored both multimodal and modality-specific representation learning. Multimodal models are trained across mixtures of imaging modalities~\cite{wu2025vision3d,xu2025generalizable3d}, enabling support for diverse image types and downstream tasks. In contrast, modality-specific models focus on a single imaging domain, including X-ray~\cite{Singla2025MultiAX, Yao2024EVAXAF}, MR~\cite{Yang2025DecipherMRAV,Dong2025MRICOREAF}, and CT~\cite{Zhu20253DFM}, often yielding stronger representations for their target modality. These models have also been adapted to specialized downstream tasks, including disease classification~\cite{takbrain26}, segmentation~\cite{Wang2023SAMMed3DTG,Zhang2024AFM}, and retrieval~\cite{Yang2025DecipherMRAV}.

However, existing medical foundation models remain limited in several important ways. Some focus on narrow anatomical regions~\cite{Cox2024BrainSegFounderT3,Barbano2024AnatomicalFM,Wang2025TriadVF}, others are optimized primarily for a single task such as segmentation~\cite{Wang2025SAMRISA,Wasserthal2022TotalSegmentatorRS}, and some operate on two-dimensional slices rather than full three-dimensional volumes~\cite{Dong2025MRICOREAF}. Models trained jointly across CT, X-ray, and MR may also generalize poorly to MR, where acquisition heterogeneity is high and large-scale labeled data are scarce.
Moreover, many existing models learn image-level representations~\cite{he2025omnimri,Dong2025MRICOREAF,Yang2025DecipherMRAV}, in some cases aligned with radiology reports, that lack explicit anatomy-level structure, making it difficult to distinguish normal from abnormal findings within a specific organ or region.

\begin{figure}[t]
\centering
\includegraphics{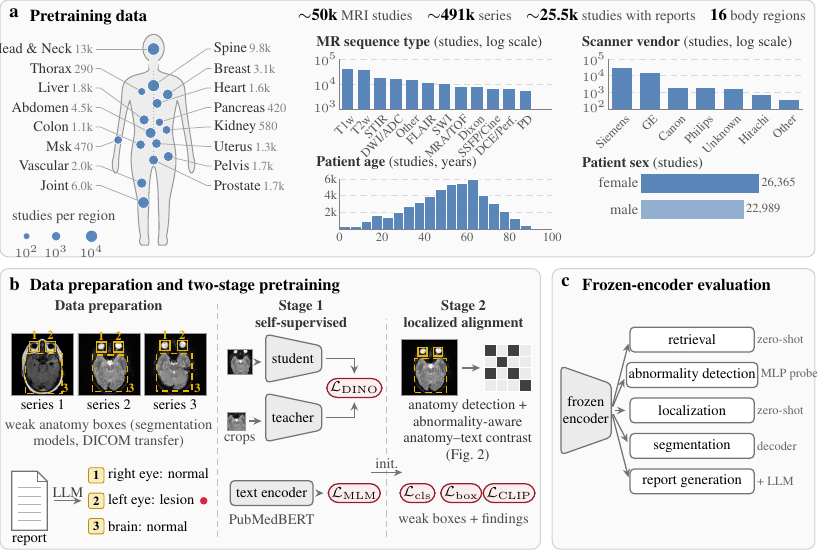}
\caption{{Overview. \textbf{(a)} Pretraining data: {MRI studies per body region (bubble size on a log scale), MR sequence types, scanner vendors, and patient age and sex (all counted per study).} \textbf{(b)} Weak anatomy boxes from segmentation models and anatomy-level findings extracted from reports by an LLM supervise a two-stage pretraining: self-supervised image (DINO student--teacher on 3D crops) and text (masked language modeling) encoders (stage~1), then localized, abnormality-aware image--text alignment (stage~2, Fig.~\ref{fig:alignment}). \textbf{(c)} Evaluation with a frozen encoder: retrieval and localization zero-shot; abnormality detection, segmentation and report generation with lightweight task-specific modules.}}
\label{fig:overview}
\label{fig:data_distribution}
\end{figure}

To address these issues, we introduce A\textsuperscript{4}-MR, an anatomy-aware multimodal pretrained model for 3D MR imaging. Our approach uses a two-stage training strategy designed to support broad generalization across MR tasks{~(Fig.~\ref{fig:overview})}.
In the first stage, self-supervised vision learning with a teacher--student framework enables the model to learn from a large-scale MR pretraining dataset. In the second stage, the model learns anatomy-targeted representations from bootstrapped
labels while aligning image and text embeddings at both the image and anatomy levels. This design enables A\textsuperscript{4}-MR to support anatomy-level localization, multimodal retrieval, and abnormality-aware representation learning, leading to improved
performance over state-of-the-art foundation models on tasks such as triaging.

\section{Related Work}
\label{sec:related_work}

Decipher-MR~\cite{Yang2025DecipherMRAV} and OmniMRI~\cite{he2025omnimri} are 3D MRI vision-language foundation models trained on large corpora of MRI series with associated radiology reports, spanning diverse anatomical regions, pathologies, and acquisition protocols. Both approaches use multi-stage training for visual and language encoders, also referred to as towers: a self-supervised vision model first learns MR image representations, followed by a second stage that aligns radiology reports with MR images. Decipher-MR uses cross-modal training in the second stage as a form of weak supervision for MR representation learning, whereas OmniMRI freezes the visual tower, aligns the text encoder, and later performs instruction tuning with additional annotated supervision.

These approaches enable general-purpose representation learning and perform strongly across retrieval, classification, and segmentation tasks compared with task-specific models. Inspired by this line of work, we adopt a multi-stage pretraining strategy in which free-form radiology reports provide weak supervision, and we further extend the representation granularity to the organ level.

MRI-CORE~\cite{Dong2025MRICOREAF} is a vision foundation model for MRI, pretrained in a self-supervised fashion on over 100k MRI volumes across 18 body regions. The model uses DINOv2 as its self-supervised training objective, combining global and image-level objectives in a student--teacher framework, and initializes its vision backbone from SAM-pretrained weights. MRI-CORE demonstrates strong performance in segmentation across body regions and in zero-shot classification tasks. However, the absence of a language component limits it to vision-only tasks and restricts its utility for applications such as retrieval and grounding. Its coverage of anatomical diversity and pathological variation is also narrower than our training corpus, motivating our approach of training natively with paired text reports across a broader range of anatomies and clinical findings.

X-DETR~\cite{Cai2022XDETRAV} is a versatile end-to-end architecture for instance-wise vision-language tasks, designed to efficiently align free-form language with individual object instances rather than whole images. Based on DETR~\cite{Carion2020EndtoEndOD} and MDETR~\cite{Kamath2021MDETRM}, it uses a visual object detector, a language encoder, and a lightweight vision-language alignment module that bridges the two modalities through an efficient dot-product operation. We adapt this idea to our MR model by aligning radiological findings with localized anatomical regions. We further extend this logic to capture the state~(e.g., healthy or abnormal) of each anatomical region in a contrastive pretraining setup.

\section{Methodology}

\subsection{Pretraining Data}
\label{sec:data}

To improve generalization to real-world settings, we compile a diverse dataset of MRI studies collected across multiple hardware platforms and geographic locations. These studies cover 16 body regions and a broad range of sequence types, scanner vendors and patient demographics (Fig.~\ref{fig:overview}a). We rigorously preprocessed the data to ensure quality, excluding scans with non-uniform spacing, missing slices, low resolution, excessive noise, or too few slices. After preprocessing, the dataset consisted of ~50k studies comprising ~491k MRI series, with reports available for ~25.5k studies. Each study included several MRI series spanning multiple protocols. We split the dataset into training and evaluation sets, using 85\% of studies for pretraining and 15\% for downstream validation. Half of the pretraining set was used for first-stage pretraining, and the full training set was used for second-stage pretraining.

\subsection{Pretraining Data Preparation}
Medical annotation is a laborious and costly process. To provide supervision at scale, we weakly annotate MR images and their corresponding radiology reports using existing models. In order to provide localization supervision, we applied standard MRI processing techniques (i.e., intensity normalization and foreground cropping; details in the Supplementary Material) and utilized an open-source segmentation model~\cite{Wasserthal2022TotalSegmentatorRS} together with proprietary custom-built deep learning (DL) segmentation models to delineate anatomical regions. We utilized DICOM header information to translate mask coordinates across different protocols in the absence of a DL model for a given sequence, as a means of weak supervision. Finally, we converted the resulting delineation masks into bounding-box coordinates to capture anatomy location. Full preprocessing details (foreground cropping, intensity scaling, and crop sizes for each pretraining stage) are provided in the Supplementary Material (Section~6.2).

To better support cross-modal training and localization, we standardized free-form radiology reports using an LLM (Claude 3.7 Sonnet). We instructed the LLM to extract medical anatomies, information describing the health status of each anatomy, and corresponding supporting evidence. Finally, we manually reviewed a subset of processed reports to check for hallucinations and ensure output consistency. A list of anatomies and additional details on anatomy-text pairing are provided in the Supplementary Material (Section~6.6).

\subsection{Pretraining}

We implement a multi stage pretraining strategy following earlier work~\cite{Yang2025DecipherMRAV, he2025omnimri, Dong2025MRICOREAF}. The first stage consists of pretraining of the image and text encoders via self-supervised methods. Unlike earlier work that focuses on image level representation, in the second round, we extend multimodal pretraining with anatomy localization where we train the encoder to identify image anatomies and align the image and text encoder embedding spaces. For the text encoder, we start with a pretrained PubMedBERT~\cite{Gu2020DomainSpecificLM} model trained on biomedical text. We then train this model on our MR reports using a masked language modeling loss. For the image encoder, we use the DINOv2 pretraining strategy of a student-teacher network trained via self supervision. The student network learns to match the teacher's representation of the MR scans at the image and patch level.

Here, our goals are to train a pretrained model that can natively do anatomy level localization, learn embedding representations at the image and anatomy level, and make the anatomy-level alignment abnormality-aware, so that normal and abnormal occurrences of an organ are treated/represented differently.

For the first goal, we leverage a DETR style decoder trained on weakly annotated data to classify and learn bounding boxes for our predefined list of anatomies. For our second goal, we use the CLS token of our encoder and leverage the fact that DETR provides output embeddings for each identified anatomy. We pass these through a projection layer and use a contrastive loss to align the anatomy with the corresponding text representation. In addition, we also align the CLS token of the encoder with the summary statement using the same text encoder. For our final goal, we leverage abnormality information from the reports to structure the contrastive targets so that the model learns to separate normal and abnormal anatomies corresponding to the same organ. Each of these is described in detail in the following sections.

\subsection{Model Architecture}
\label{sec:model-arch}
\begin{figure}[htbp]
\centering
\includegraphics{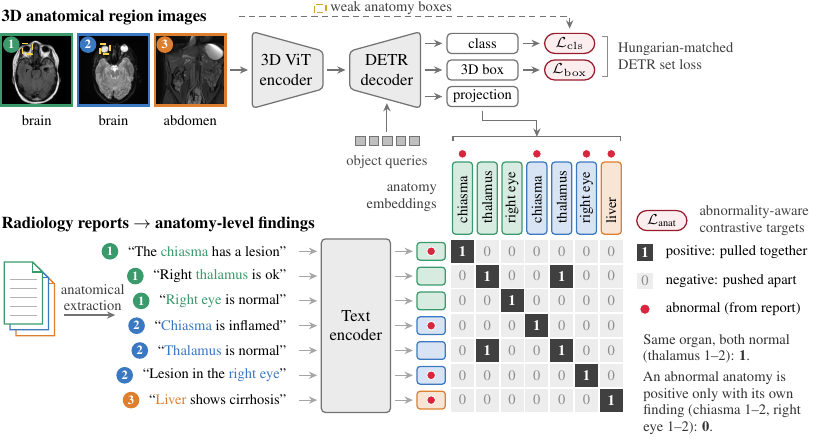}
\caption{{Cross-modal localized alignment. A 3D ViT encoder and a DETR decoder detect anatomies in each MR volume. Each decoder output feeds three heads: class and 3D box, trained with the Hungarian-matched DETR set loss, and a projection that gives one embedding per detected anatomy. Similarly, report findings are extracted per anatomy from the radiology reports and are encoded by a text encoder.
The abnormality-aware contrastive loss $\mathcal{L}_{\text{anat}}$ (Eq.~\ref{eq:anat-loss}) uses pairwise target matrix where the same organ in two images is a positive pair only if both are normal (thalamus); an abnormal anatomy is positive only with its own finding (chiasma, right eye). The image-level CLS--text contrastive loss is omitted for clarity.~(best in color)}}
\label{fig:alignment}
\end{figure}
{Fig.~\ref{fig:alignment} gives an overview of the second-stage architecture and losses.}
Augmentation details for both pretraining stages and study-level, organ-aware data sampling strategy are provided in the Supplementary Material~(Section~6.3 and Section~6.4 respectively).

We employ a 3D Vision Transformer (ViT-Base) as our vision encoder similar to~\cite{Dong2025MRICOREAF, tak2024foundation, Yang2025DecipherMRAV, Wang2023SAMMed3DTG}. We use an $8 \times 8 \times 8$ cubic non-overlapping patch size. The model consists of a projection layer that operates on flattened patches to generate patch embeddings. A linear position embedding interpolation allows handling of different image sizes during downstream and inference tasks. Our model follows the rest of the standard ViT-B architecture~\cite{dosovitskiy2021an}.
Our decoder used for localization is based on DETR~\cite{Carion2020EndtoEndOD}. We use 200 object queries, 126 classes including the no object class and a 6 layer transformer, that consist of self attention, cross attention and an MLP for bounding box regression and classification. In addition, we use the output object embedding from the decoder projected through a linear layer to align the anatomy representation with text embeddings. Section~\ref{sec:detr} provides a more detailed description of the DETR architecture.
We use a BERT style encoder initialized from PubMedBERT, which was trained on biomedical text data. We train this encoder via a projection layer to align with the CLS token of the ViT-based teacher encoder, and object query embeddings of the DETR decoder. {Both alignments use a sigmoid contrastive loss over a multi-positive target matrix (Section~\ref{sec:auramr-alignment}), adapted from the image--text contrastive learning of CLIP~\cite{pmlr-v139-radford21a}.}

\subsubsection{DETR for bounding box detection}
\label{sec:detr}
We use the Detection Transformer architecture proposed by~\cite{Carion2020EndtoEndOD} to build our anatomy localizer. As a refresher, DETR operates as follows.

It consists of a backbone such as ResNet, a transformer encoder, and a transformer decoder. We get rid of the ResNet and directly feed images into a ViT based transformer encoder. The output tokens from the transformer are Z. The transformer decoder takes learnable input queries Q, and consists of L stacked decoder blocks. Each block consists of a self attention layer followed by a cross attention layer. The latter makes the object queries interact with the encoder tokens Z. The output embeddings from each decoder layer are then used to train an MLP classifier to classify each embedding into a predefined list of classes or a no object found class. A regression head then generates the box coordinates for classes other than no object found. In addition, we add a contrastive projection head to the output embeddings of the last layer, which is then used to align anatomy text representations. We describe the working of this in Section~\ref{sec:auramr-alignment}.

Following DETR, we employ a fixed set of $N$ object queries, where $N$ is chosen to exceed the maximum number of anatomical structures present in an MRI scan. The ground-truth set $\hat{y}$ is correspondingly padded with $\varnothing$ (no-object) labels to match. This formulation casts detection as a set prediction problem, which we solve via bipartite matching using the Hungarian algorithm.

Let $\sigma$ denote the optimal permutation over $N$ elements obtained from the Hungarian matching. The training objective is then,

\begin{equation}
\mathcal{L}_{\mathrm{loss}}(y, \hat{y}) =
\sum_{i=1}^{N} \left[
- \log \hat{p}_{\sigma(i)}(c_i)
+ \mathbbm{1}_{\{c_i \neq \varnothing\}} \, \mathcal{L}_{\mathrm{box}}(b_i, \hat{b}_{\sigma(i)})
\right],
\end{equation}

where $c_i$ and $b_i$ denote the class label and bounding box of the $i$-th ground-truth element, respectively, and $\hat{p}_{\sigma(i)}(c_i)$ and $\hat{b}_{\sigma(i)}$ denote the predicted class probability and bounding box for the matched prediction $\sigma(i)$. The indicator function $\mathbbm{1}_{\{c_i \neq \varnothing\}}$ ensures that the bounding box loss is applied only to non-empty targets.

The bounding box loss is defined as a weighted combination of an L1 loss and GIoU loss following DETR. We provide the equation here for completeness

\begin{equation}
\mathcal{L}_{\mathrm{box}}(b_i, \hat{b}_{\sigma(i)}) =
\lambda_{\mathrm{L1}} \, \mathcal{L}_{\mathrm{L1}}(b_i, \hat{b}_{\sigma(i)})
+ \lambda_{giou} \, \mathcal{L}_{\mathrm{giou}}(b_i, \hat{b}_{\sigma(i)})
\end{equation}

where $\lambda_{\mathrm{L1}}, \lambda_{giou} \in \mathbb{R}$ are weighting coefficients.

In addition to the primary matching loss, we apply auxiliary supervision at each decoder layer, following DETR. Furthermore, we introduce two additional objectives. First, we incorporate an image--text contrastive loss between the CLS tokens of the vision encoder (ViT) and the text encoder, using the sigmoid formulation of Eq.~\ref{eq:anat-loss} with $M_{jk}=\mathbbm{1}[s_j=s_k]$. Second, we propose an anatomy-level contrastive loss that aligns the CLS token of the text encoder with the object queries from the final decoder layer. 

\subsubsection{\texorpdfstring{A\textsuperscript{4}-MR}{A4-MR} for anatomy-level multimodal alignment with abnormality awareness}
\label{sec:auramr-alignment}
{%
Image-level vision-language pretraining treats a scan and its report as one positive pair. Within an anatomy, however, normal and abnormal appearances behave differently: healthy instances of an anatomy look alike across patients, whereas abnormal findings are heterogeneous (one abnormal liver may be cirrhotic, another may contain a metastasis). We therefore construct the contrastive targets so that (i)~normal occurrences of the same anatomy form one shared positive class across images, (ii)~an abnormal occurrence is positive only with the findings for that anatomy in its own image, and (iii)~normal and abnormal occurrences of the same anatomy are explicit negatives. Healthy anatomy is pulled towards an anatomy-specific prototype, while abnormal anatomy is kept apart from it and from other patients' abnormalities (Fig.~\ref{fig:alignment}).

For each image $b$ in a batch of $B$ images, the DETR decoder outputs $Q=200$ query embeddings, which the projection head $f_{\text{clip}}$ maps to $\ell_2$-normalized vectors of dimension $H_c=512$. The LLM splits the report of image $b$ into anatomy-level findings (Supplementary Section~6.5), each with an organ label $o \in \{1,\dots,C\}$ and an abnormality flag $a \in \{0,1\}$. Each finding is paired with the query that the Hungarian matcher assigns to the ground-truth box of the same organ in image $b$; findings whose organ has no box in the image are discarded. This yields $M$ pairs $\{(\mathbf{v}_j, \mathbf{t}_j, o_j, s_j, a_j)\}_{j=1}^{M}$, where $\mathbf{v}_j$ is the visual anatomy embedding, $\mathbf{t}_j$ the $\ell_2$-normalized text embedding and $s_j$ the source image.

A report may mention an organ several times (e.g., ``the liver is enlarged'' and ``no focal liver lesion''). Since each image has a single embedding per organ, its state must be consistent, so we mark the organ abnormal if any of its findings is abnormal:
\begin{equation}
  \tilde{a}_j = \max_{k:\; s_k = s_j,\; o_k = o_j} a_k .
\end{equation}

The target for visual embedding $j$ and text $k$ is
\begin{equation}
  \label{eq:anat-targets}
  M_{jk} = \mathbbm{1}[o_j = o_k]\Big(
  \underbrace{(1-\tilde{a}_j)(1-\tilde{a}_k)}_{\text{both normal}}
  + \underbrace{\tilde{a}_j\,\tilde{a}_k\,\mathbbm{1}[s_j = s_k]}_{\text{same abnormal organ, same image}}
  \Big).
\end{equation}
Different organs, and normal--abnormal occurrences of the same organ, are negatives. Fig.~\ref{fig:alignment} shows the resulting matrix for three images: the two normal thalami are positives, whereas the two abnormal chiasmata, which come from different images, are negatives.

Because a row of $M$ can contain several positives, we use a sigmoid (binary cross-entropy) contrastive loss instead of the softmax CLIP loss:
\begin{equation}
  \label{eq:anat-loss}
  \mathcal{L}_{\text{anat}} = -\frac{1}{M^2}\sum_{j,k}\Big[M_{jk}\log\sigma(z_{jk}) + (1-M_{jk})\log\big(1-\sigma(z_{jk})\big)\Big],
  \qquad z_{jk} = \tau\, \mathbf{v}_j^{\top}\mathbf{t}_k ,
\end{equation}
where $\tau$ is a learnable scale initialized to $1/0.07$ and clamped at 100. As $M$ is symmetric, this single term covers both the image-to-text and text-to-image directions. The image-level loss between the CLS token and the full report uses the same formulation with $M_{jk} = \mathbbm{1}[s_j = s_k]$. For the ablation A\textsuperscript{4}-MR-CLIP, the anatomy--text loss uses this study-level target as well, $M^{\text{CLIP}}_{jk} = \mathbbm{1}[s_j = s_k]$: every anatomy embedding of an image is a positive for every finding of that image, regardless of anatomy or abnormality, and all pairs from other images are negatives. It therefore drops both the anatomy term and the normal/abnormal rule of Eq.~\ref{eq:anat-targets}, while the loss, the detector and the data are unchanged.

Abnormal occurrences of the same anatomy in different patients are treated as negatives even when they share a diagnosis, because report-derived flags indicate \emph{that} an organ is abnormal but not \emph{how}. We discuss these false negatives in Section~\ref{sec:limitations}.
}%

\section{Results}
\label{sec:results}
We compare our model to the following models:
\begin{itemize}[leftmargin=0.5cm]
  \item Vista3D CT-MR~\cite{he2024vista3d}, a multimodal 3D foundation model designed to jointly learn from CT and MR volumes for robust cross-modality medical image segmentation.
  \item MedSigLIP~\cite{Sellergren2025MedGemmaTR}, a medical adaptation of the SigLIP~\cite{Zhai2023SigmoidLF} vision-language model that aligns clinical images with text for improved multimodal understanding and retrieval.
  \item BiomedParse~\cite{zhao2024biomedparse}, a large-scale biomedical parsing model that unifies segmentation and structured understanding across diverse medical imaging tasks and anatomies.
  \item MRI-CORE~\cite{Dong2025MRICOREAF}, a foundation model trained specifically on MRI data to capture modality-specific representations for downstream tasks such as segmentation and diagnosis.
  \item Decipher-MR~\cite{Yang2025DecipherMRAV}, where a 3D volumetric image level contrastive objective is utilized to align visual and text towers.
  \item MedSAM2~\cite{medsam2}, a promptable medical segmentation foundation model that propagates 2D box prompts through a 3D volume.
  \item nnUNet3D~\cite{Isensee2020nnUNetAS}, a deep learning-based segmentation method that automatically configures itself, including preprocessing, network architecture, training and post-processing, for any new task.
  \item MedNeXt~\cite{Roy2023MedNeXtTS}, a ConvNeXt-based 3D segmentation network trained end-to-end.
  \item VNet~\cite{Milletar2016VNetFC}, a 3D convolutional volumetric segmentation model, originally proposed for prostate MRI.
  \item BiomedCLIP~\cite{Zhang2023BiomedCLIPAM}, a multimodal vision-language foundation model trained on image-text pairs from biomedical scientific articles.
  \item MedImageInsight~\cite{Codella2024MedImageInsightAO}, a medical imaging embedding model trained on images and text.
  \item Qwen2.5-VL~\cite{Bai2025Qwen25VLTR} (3B, 32B) and MedGemma-1.5-4B~\cite{Sellergren2025MedGemmaTR}, vision-language models used zero-shot for report generation.
  \item A\textsuperscript{4}-MR-CLIP, an ablation of A\textsuperscript{4}-MR trained with the same setup but in which the anatomy--text loss uses the study-level target of the image-level loss ($M_{jk}=\mathbbm{1}[s_j=s_k]$, Section~\ref{sec:auramr-alignment}): each anatomy is paired with all findings of its own image, without anatomy- or abnormality-conditioned positives.
\end{itemize}

\paragraph{Datasets.}
{We evaluate on public and proprietary datasets covering retrieval, classification, segmentation, localization and report generation.}
\begin{itemize}[leftmargin=0.5cm]
  \item {{Retrieval set}, {an external dataset not used in pretraining: 380 MRI studies with radiology reports (3{,}346 series) across nine body regions}. Metrics: Precision @K and mean average precision~(mAP).}
  \item {{CirrMRI600+}~\cite{jha2025large}, 310 T1 and 318 T2 abdominal MRI volumes of cirrhotic livers and 55 healthy volumes per sequence, with liver masks.}
  {Official train, validation and test splits exist for the cirrhotic scans only. For abnormality detection, we keep the official cirrhotic test set (62 scans), merge the official training and validation scans into one development set (566 scans), and cross-validate the healthy scans, which have no official split: the 55 healthy subjects are divided into five subject-grouped folds, each fold joining the test set in turn, for two random partitions. For segmentation, we evaluate on a held-out test set.} Metrics: AUROC (classification), Dice (segmentation).
  \item {{ADNI}~\cite{petersen2010alzheimer}, T1-weighted brain MRI of 1{,}731 subjects, one scan each (339 AD, 871 MCI and 521 cognitively normal (CN); all scans sagittal). We follow the Decipher-MR evaluation protocol~\cite{Yang2025DecipherMRAV}: 50 random splits (about 80\% training, 10\% validation and 10\% test, stratified by diagnosis), the same for every model; for the 10\% budget, the training set is subsampled with the same stratification. Tasks: AD vs CN and MCI vs CN. Metric: AUROC.}
  \item {{AMOS22 (MR)}~\cite{ji2022amos}, abdominal MRI with 13 annotated organs.
   We use the official split (40 training, 20 validation volumes). Localization is evaluated on the 20 validation volumes (13 organs present in MR, 259 image--organ pairs,
   excluding 12 pairs with unreliable ground-truth boxes)\footnote{ labels with stray voxels that enlarge the box more than 1.3-fold (7) and organs thinner than one 8-voxel patch along an axis (6, one pair meets both). The same pairs are excluded for every model; this changes scores by at most 1.4 points and does not change the ranking (Supplementary Section~7.4).}}
  \item {{Dataset A (pelvis)}, proprietary, {193} pelvic MRI volumes with 11 annotated organs that are not part of the weak pretraining labels; {140 training, 18 validation and 35 test volumes}. Metrics: Mean Dice.}
  \item {{Dataset B (prostate)}, proprietary, around 1{,}000 prostate MRI studies, each with 3 series (T2W, ADC, DWI) and a radiology report. The fairness analysis (Section~\ref{sec:fairness}) uses the 718 studies with biopsy- or prostatectomy-confirmed ISUP grade. Report generation (Section~\ref{sec:report_gen}) uses 250 studies for training and 50 for checkpoint selection, and a balanced test set of 60 studies drawn from the remaining 700 (one per patient; 30 PI-RADS $\le$2 and 10 each of PI-RADS 3, 4 and 5, half GE and half Siemens in every group) (label extraction and checks: Supplementary Sec.~8.1).}
  \item {{Dataset C (knee)}, proprietary, 1,357 training, 60 validation and 50 test 3D knee MRI volumes with femoral, tibial and patellar cartilage {and meniscus} labels. Metrics: Mean Dice.}
\end{itemize}
{Additional ablation results (classification, age regression) are reported in the Supplementary Material.}

\FloatBarrier
\subsection{Cross Modal Retrieval}
We evaluated A\textsuperscript{4}-MR's zero-shot cross-modal retrieval in both directions (text-to-image and image-to-text) on an external dataset, not used in pretraining, with reports and MRIs across nine body regions, comparing it against other biomedical vision-language models. Reports were first structured by an LLM into one section per anatomy plus a conclusion (Supplementary Sec.~6.5). We use two kinds of text queries: the entire structured report (all sections joined; one query per study) and each report section separately (mostly organ-level descriptions; one query per section). A retrieval counts as correct at $N$ if all top-$N$ retrieved items belong to the query's body region; we report this at $N = 1, 3, 5$ together with mean average precision (mAP) over body-region relevance.

As shown in Tables~\ref{tab:img2txt_combined} and \ref{tab:txt2img_combined}, A\textsuperscript{4}-MR achieves overall better performance across image-to-text and text-to-image retrieval tasks, consistently outperforming other cross-modal models such as Decipher-MR, BiomedCLIP, and MedImageInsight. In Table~\ref{tab:img2txt_combined}, A\textsuperscript{4}-MR obtains the highest scores for all Entire Report image-to-text retrieval metrics and achieves the best performance in three of four Organ Description metrics, with only a marginal decrease in @1 compared to the Decipher-MR (85.42 vs.\ 86.60). Similarly, Table~\ref{tab:txt2img_combined} shows that A\textsuperscript{4}-MR demonstrates the strongest overall text-to-image retrieval performance across the Entire Report and Organ Description settings, with the exception of Entire Report @1, where A\textsuperscript{4}-MR-CLIP achieves the highest score. The largest gains are observed in mAP and higher-rank retrieval metrics (@3 and @5), indicating improved ranking quality and stronger cross-modal alignment rather than only better top-1 retrieval. Overall, these results demonstrate that A\textsuperscript{4}-MR effectively captures both global report-level semantics and fine-grained organ-specific information, leading to more robust and accurate medical image-text retrieval.

\begin{table}[ht]
\centering
\caption{Image-to-Text Performance on body region retrieval task. {Bold: best; underlined: second best.}}
\label{tab:img2txt_combined}
\begin{tabular}{lcccccccc}
\toprule
& \multicolumn{4}{c}{Entire Report} & \multicolumn{4}{c}{Organ Descriptions} \\
\cmidrule(lr){2-5} \cmidrule(lr){6-9}
Model & @1 & @3 & @5 & mAP & @1 & @3 & @5 & mAP \\
\midrule
A\textsuperscript{4}-MR (ours)             & \textbf{93.27} & \textbf{86.52} & \textbf{84.13} & \textbf{90.13} & \underline{85.42} & \textbf{74.51} & \textbf{68.74} & \textbf{77.58} \\
Decipher-MR      & \underline{90.90} & 79.60 & 71.90 & 77.80 & \textbf{86.60} & \underline{72.30} & \underline{62.80} & 66.70 \\
BiomedCLIP       & 81.90 & 69.80 & 64.40 & 70.90 & 66.40 & 48.80 & 41.00 & 51.70 \\
MedImageInsight  & 78.00 & 65.10 & 58.10 & 71.70 & 73.00 & 57.30 & 50.50 & 60.70 \\
A\textsuperscript{4}-MR-CLIP (ours)   & 88.25 & \underline{82.49} & \underline{78.87} & \underline{84.38} & 81.47 & 68.71 & 60.76 & \underline{70.66} \\
\bottomrule
\end{tabular}
\end{table}

\begin{table}[ht]
\centering
\caption{Text-to-Image Retrieval Performance. {Bold: best; underlined: second best.}}
\label{tab:txt2img_combined}
\begin{tabular}{lcccccccc}
\toprule
& \multicolumn{4}{c}{Entire Report} & \multicolumn{4}{c}{Organ Descriptions} \\
\cmidrule(lr){2-5} \cmidrule(lr){6-9}
Model & @1 & @3 & @5 & mAP & @1 & @3 & @5 & mAP \\
\midrule
A\textsuperscript{4}-MR (ours)             & \underline{96.49} & \textbf{94.60} & \textbf{92.70} & \textbf{88.32} & \textbf{86.93} & \textbf{82.58} & \textbf{79.53} & \textbf{78.42} \\
Decipher-MR      & 94.60 & 91.40 & \underline{89.70} & 78.10 & \underline{84.50} & \underline{78.80} & \underline{75.10} & \underline{73.80} \\
BiomedCLIP       & 83.50 & 72.40 & 68.10 & 75.50 & 66.00 & 52.70 & 47.40 & 62.40 \\
MedImageInsight  & 91.90 & 81.10 & 74.90 & 77.20 & 73.90 & 63.70 & 56.90 & 66.80 \\
A\textsuperscript{4}-MR-CLIP (ours)   & \textbf{97.30} & \underline{91.89} & 87.84 & \underline{81.16} & 79.92 & 71.28 & 65.65 & 72.24 \\
\bottomrule
\end{tabular}
\end{table}
\FloatBarrier
\subsection{Classification}
\subsubsection{Improved abnormality detection}

\begin{figure}[htbp]
\centering
\includegraphics[width=0.95\textwidth]{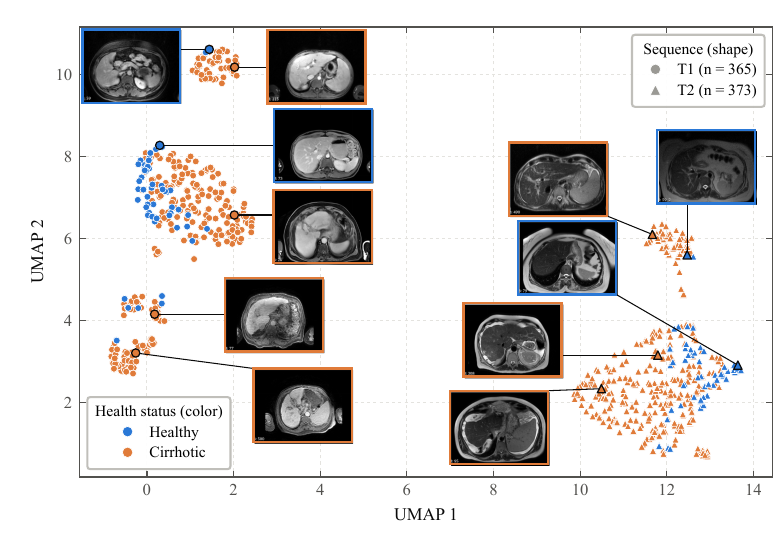}
\caption{UMAP~\cite{McInnes2018UMAP} plot of the full CirrMRI600+ dataset with A\textsuperscript{4}-MR (ours) model embeddings. Colors indicate the health status of the sample. Shapes indicate the MRI pulse sequence of the image that we have found to be an important factor for the model.}
\label{fig:cirrmri_umap}
\end{figure}

{We test the model's representation capabilities on disease detection using the CirrMRI600+ dataset~(Fig.~\ref{fig:cirrmri_umap}). 
On frozen embeddings, we train an MLP probe with two hidden layers (768 units each), a single
learning rate of $10^{-5}$ for all layers (chosen once by mean validation AUROC over all models) and a fixed
budget of 10k steps; the last checkpoint is evaluated, and results are averaged over 5 folds $\times$ 2
partitions $\times$ 3 seeds (see Supplementary for details and results on the official split). Embeddings come
from a single forward pass on the whole volume resized to $384 \times 384 \times 48$ in its stored orientation,
which matches the pretraining layout, and the same probe and splits are used for every model and representation. To measure data efficiency, we train on 10\%, 25\%, 50\% and 100\% of the development set while keeping the class prevalence. Additional classification results are provided in the Subgroup Analysis Section~\ref{sec:subgroup}.}

\begin{table}[htbp]
\centering
\caption{{Comparison of models on CirrMRI600+ abnormality detection across different data budgets in training (AUROC, mean $\pm$ std over 30 probes: 5 healthy-subject folds $\times$ 2 partitions $\times$ 3 seeds, evaluated on the official cirrhotic test set plus the held-out healthy subjects).} {Bold: best; underlined: second best.}}
\label{tab:cirrmri_attribute_detection}
\begin{tabular}{lcccc}
\toprule
 & \multicolumn{4}{c}{Abnormality detection (AUROC)} \\
\cmidrule(lr){2-5}
Model & 10\% & 25\% & 50\% & 100\% \\
\midrule

A\textsuperscript{4}-MR (ours) & \underline{91.6 $\pm$ 5.0} & \underline{97.1 $\pm$ 2.3} & \underline{98.3 $\pm$ 1.4} & \underline{98.9 $\pm$ 0.8} \\
A\textsuperscript{4}-MR-box (ours) & \textbf{95.3 $\pm$ 3.6} & \textbf{97.9 $\pm$ 1.9} & \textbf{98.8 $\pm$ 1.0} & \textbf{99.0 $\pm$ 1.0} \\
Decipher-MR & 85.9 $\pm$ 5.2 & 92.4 $\pm$ 3.5 & 96.0 $\pm$ 2.4 & 97.0 $\pm$ 2.0 \\
MedSigLIP & 89.9 $\pm$ 3.9 & 94.0 $\pm$ 2.6 & 95.3 $\pm$ 2.0 & 96.2 $\pm$ 1.8 \\
MedGemma & 88.1 $\pm$ 3.3 & 91.5 $\pm$ 3.3 & 93.7 $\pm$ 2.1 & 95.1 $\pm$ 1.6 \\
MedImageInsight & 88.1 $\pm$ 3.8 & 92.3 $\pm$ 3.0 & 93.7 $\pm$ 3.3 & 94.6 $\pm$ 2.8 \\
MRI-CORE & 85.6 $\pm$ 4.5 & 90.9 $\pm$ 3.1 & 94.8 $\pm$ 3.0 & 96.3 $\pm$ 2.6 \\
BiomedParse & 79.5 $\pm$ 5.2 & 86.1 $\pm$ 3.7 & 89.4 $\pm$ 4.5 & 91.0 $\pm$ 4.2 \\
Vista3D CT-MR & 79.9 $\pm$ 5.5 & 84.9 $\pm$ 3.7 & 88.3 $\pm$ 4.7 & 90.7 $\pm$ 3.2 \\

\bottomrule
\end{tabular}
\end{table}

{Table~\ref{tab:cirrmri_attribute_detection} shows the results of the abnormality detection task. A\textsuperscript{4}-MR gives the highest AUROC at every data budget. 
We also looked into localized representation capabilities, e.g. given an anatomy detection result, building a representation from specific anatomy by pooling the corresponding backbone patch tokens~(A\textsuperscript{4}-MR-box; Tab.~\ref{tab:cirrmri_ablations}). Liver-box representation raises the AUROC to 95.3, 97.9, 98.8 and 99.0, and is significantly better than every baseline at every budget, including the 10\% budget (+5.4 AUROC over MedSigLIP, the strongest baseline there). 

\begin{table}[!ht]
\centering
\caption{Ablations on how the image is turned into one embedding, on CirrMRI600+ (cirrhosis vs healthy) and ADNI (AD vs CN and MCI vs CN), with 10\%, 25\%, 50\% and 100\% of the training data on CirrMRI600+ and 10\% and 100\% on ADNI (AUROC $\times 100$, mean $\pm$ std). 
Patch mean averages all 13{,}824 patch tokens. Box pool averages the backbone patch tokens inside the predicted target-organ box, in the same forward pass; the size-matched random box pools the same number of tokens from a box of the same size placed at a random foreground position. Crop CLS runs the model a second time on the image cropped to the box. A\textsuperscript{4}-MR uses abnormality-conditioned pairwise anatomy--text affinity targets; A\textsuperscript{4}-MR-CLIP uses study-level targets for the anatomies (each anatomy paired with all findings of its own image; no organ- or abnormality-conditioned positives). GT crop CLS uses the ground-truth liver mask (upper bound, CirrMRI600+ only) and is not ranked. Bold: best; underlined: second best.} \label{tab:cirrmri_ablations}
\setlength{\tabcolsep}{3.5pt}
\resizebox{\textwidth}{!}{%
\begin{tabular}{lcccccccc}
\toprule
 & \multicolumn{4}{c}{CirrMRI600+} & \multicolumn{2}{c}{ADNI: AD vs CN} & \multicolumn{2}{c}{ADNI: MCI vs CN} \\
\cmidrule(lr){2-5}\cmidrule(lr){6-7}\cmidrule(lr){8-9}
Representation & 10\% & 25\% & 50\% & 100\% & 10\% & 100\% & 10\% & 100\% \\
\midrule
\multicolumn{9}{l}{\textit{Whole image}} \\
CLS & 91.6 $\pm$ 5.0 & \underline{97.1 $\pm$ 2.3} & \underline{98.3 $\pm$ 1.4} & \underline{98.9 $\pm$ 0.8} & 65.8 $\pm$ 7.4 & 80.5 $\pm$ 5.0 & 52.5 $\pm$ 5.9 & 58.9 $\pm$ 5.1 \\
Patch mean (all tokens) & 90.3 $\pm$ 4.8 & 95.4 $\pm$ 2.8 & 97.2 $\pm$ 2.0 & 97.6 $\pm$ 1.6 & 62.6 $\pm$ 8.2 & 79.3 $\pm$ 6.6 & 51.5 $\pm$ 5.6 & 59.0 $\pm$ 5.3 \\
A\textsuperscript{4}-MR-CLIP: CLS & 85.1 $\pm$ 5.3 & 93.7 $\pm$ 2.4 & 96.3 $\pm$ 1.8 & 97.1 $\pm$ 0.9 & 64.2 $\pm$ 8.5 & 80.3 $\pm$ 4.6 & 52.3 $\pm$ 5.6 & 57.4 $\pm$ 5.2 \\
A\textsuperscript{4}-MR-CLIP: patch mean & 83.2 $\pm$ 5.4 & 90.3 $\pm$ 3.7 & 93.5 $\pm$ 3.2 & 94.7 $\pm$ 2.7 & 60.2 $\pm$ 6.7 & 77.3 $\pm$ 6.4 & 51.4 $\pm$ 5.6 & 55.6 $\pm$ 6.1 \\
\midrule
\multicolumn{9}{l}{\textit{Detection head (organ queries)}} \\
Mean organ & 85.2 $\pm$ 4.0 & 90.3 $\pm$ 3.5 & 92.1 $\pm$ 2.4 & 93.2 $\pm$ 2.2 & 64.6 $\pm$ 6.7 & 74.5 $\pm$ 5.6 & 49.6 $\pm$ 5.5 & 56.5 $\pm$ 5.2 \\
Target-organ query (liver / MTL) & 89.3 $\pm$ 4.7 & 93.2 $\pm$ 3.1 & 95.1 $\pm$ 2.5 & 96.1 $\pm$ 2.5 & \underline{71.4 $\pm$ 6.5} & 81.8 $\pm$ 5.0 & \underline{52.8 $\pm$ 6.2} & 60.0 $\pm$ 5.4 \\
CLS + target-organ query & \underline{91.8 $\pm$ 4.9} & 96.8 $\pm$ 2.0 & 98.0 $\pm$ 1.4 & 98.5 $\pm$ 1.2 & 71.0 $\pm$ 6.5 & 84.2 $\pm$ 4.2 & 51.8 $\pm$ 5.5 & \underline{60.6 $\pm$ 5.1} \\
\midrule
\multicolumn{9}{l}{\textit{Target-organ-localized backbone features}} \\
Target-organ box pool (liver / MTL) & \textbf{95.3 $\pm$ 3.6} & \textbf{97.9 $\pm$ 1.9} & \textbf{98.8 $\pm$ 1.0} & \textbf{99.0 $\pm$ 1.0} & \textbf{75.7 $\pm$ 7.0} & \textbf{88.1 $\pm$ 4.4} & \textbf{53.5 $\pm$ 6.8} & \textbf{63.6 $\pm$ 4.4} \\
Target-organ crop CLS & 91.2 $\pm$ 4.5 & 96.0 $\pm$ 2.7 & 97.7 $\pm$ 1.6 & 98.5 $\pm$ 1.0 & 69.6 $\pm$ 7.4 & \underline{85.4 $\pm$ 3.9} & 51.7 $\pm$ 5.7 & 59.3 $\pm$ 4.2 \\
Size-matched random box (control) & 78.1 $\pm$ 7.3 & 88.7 $\pm$ 4.2 & 93.5 $\pm$ 2.4 & 95.5 $\pm$ 1.8 & 53.4 $\pm$ 6.6 & 64.9 $\pm$ 5.6 & 50.5 $\pm$ 4.8 & 56.9 $\pm$ 5.4 \\
\textit{GT crop CLS} & 91.4 $\pm$ 5.5 & 96.2 $\pm$ 2.3 & 98.1 $\pm$ 1.2 & 98.7 $\pm$ 0.6 & -- & -- & -- & -- \\
\bottomrule
\end{tabular}}
\end{table}

{To understand where the advantage of A\textsuperscript{4}-MR comes from, we utlized different embedding extraction approaches using the same probes and protocols as in Tab.~\ref{tab:cirrmri_attribute_detection} and, for ADNI, the Decipher-MR evaluation protocol. The target organ is the liver on CirrMRI600+ and the medial temporal lobe (MTL: hippocampus, amygdala and inferior lateral ventricle, left and right) on ADNI; every ADNI target-organ row uses these six structures (query: the six queries averaged; box pool: tokens pooled per box, then averaged; crop: the union of the six boxes).

\textbf{Whole image.} The abnormality-conditioned targets improve the backbone where the abnormality labels match the task: on CirrMRI600+, the CLS token of A\textsuperscript{4}-MR-CLIP is significantly worse than that of A\textsuperscript{4}-MR at every data budget ($-6.5$, $-3.4$, $-2.1$ and $-1.8$ AUROC at 10\%, 25\%, 50\% and 100\% data), and so is its patch mean ($-7.2$ to $-3.0$). On ADNI the two models are on par (CLS within 1.6 AUROC, not significant at any budget). 

\textbf{Organ queries.} The embeddings produced by the detection head stay below the CLS token on CirrMRI600+: the liver query by $-2.3$ to $-3.9$ AUROC (significant from 25\% data) and the mean over all detected organs by $-5.7$ to $-6.8$; concatenating the CLS token with the liver query recovers the level of the CLS token but does not exceed it. On ADNI, the averaged MTL queries add to the CLS token: $+5.6$ AUROC for AD at 10\% data (significant), and the CLS token concatenated with the MTL queries improves AD detection at both budgets ($+5.3$ and $+3.8$, significant). 

\textbf{Organ-localized backbone features.} Pooling the backbone patch tokens inside the predicted organ box is the best representation in every column. It improves over the CLS token on both datasets, most clearly when labels are scarce (CirrMRI600+: $+3.7$ at 10\% data, significant, and $+0.1$ to $+0.8$ from 25\% data; ADNI AD: $+9.9$ and $+7.7$ at 10\% and 100\% data; ADNI MCI: $+4.7$ at 100\% data; all significant). Two controls show that this gain comes from \emph{where} the tokens are taken. Pooling all tokens (patch mean) is significantly worse than the box pool (CirrMRI600+: $-5.0$ to $-1.4$; ADNI: $-13.1$ and $-8.9$ for AD, $-4.6$ for MCI at 100\% data), and pooling the same number of tokens from a box of the same size at a random position (size-matched random box) is worse still (CirrMRI600+: $-17.2$ to $-3.6$; ADNI AD: $-22.2$ and $-23.2$; all significant). 
On CirrMRI600+, the predicted box is as good as the ground truth for this task, and the benefit of localization comes from pooling the backbone tokens of the organ rather than from re-encoding a cropped image.
\par}

\FloatBarrier
\subsection{Segmentation}
\label{sec:segmentation}
We evaluate whether the representations learned by A\textsuperscript{4}-MR transfer to dense prediction by training a lightweight segmentation decoder on top of the frozen vision encoder. In all experiments, the encoder weights are kept fixed and only the segmentation decoder is trained, allowing us to isolate the quality of the pretrained image representations.

We use a SegResNet-style decoder~\cite{Myronenko20183DMB} with skip connections from the A\textsuperscript{4}-MR encoder. Specifically, the final patch embeddings and intermediate embeddings from zero indexed layers 2, 5 and 8 are provided to the decoder to preserve spatial detail at multiple resolutions.

We experimented with several voxel spacings and found that smaller spacings consistently improved segmentation performance, even when the original acquisition spacing was coarser. Unless otherwise stated, we use the 25th percentile spacing for each dataset. Models are trained with a learning rate of $5 \times 10^{-4}$, cosine decay, and a minimum learning rate of $10^{-6}$. Training is performed for 60k batches per GPU across 8 GPUs with a batch size of 2. We use random cropping with a crop size of $384 \times 384 \times 48$ and additional augmentations adapted from AMOS~\cite{ji2022amos}.

We benchmark A\textsuperscript{4}-MR against alternative pretrained models on three segmentation datasets: AMOS~\cite{ji2022amos}, an internal Dataset A (pelvis), and CirrMRI600+. Dataset A is out of distribution relative to the pretraining corpus and splits are given under Datasets Section. 
Knee cartilage segmentation is analyzed separately in Section~\ref{sec:kneesegmentation}. Unless otherwise stated, all results are reported as mean Dice across segmentation classes. Extended per-organ ablations (encoder variant, curriculum augmentation, decoder warm-start) are provided in the Supplementary Material.

\begin{table}[ht]
\centering
\caption{Segmentation Dice scores (\%, higher is better). {Bold: best; underlined: second best.}}
\label{tab:amos_segmentation}
\begin{tabular}{lccc}
\toprule
Model & AMOS  & Dataset A (pelvis) & CirrMRI600+\\
\midrule
A\textsuperscript{4}-MR (ours)             & 85.3  & \underline{87.6} & \textbf{96.5}\\
Decipher-MR   & \textbf{85.7}  & 84.3 & \underline{96.3} \\
nnUNet3D & \underline{85.6}  & \textbf{88.7} & 84.8\\
\bottomrule
\end{tabular}
\end{table}

\subsubsection{Comparison Against End-to-end~(e2e) models}
\label{sec:kneesegmentation}

We use the same frozen A\textsuperscript{4}-MR encoder and the same lightweight SegResNet decoder~\cite{Myronenko20183DMB} as for the other segmentation tasks, and compare against e2e models on the KneeSeg dataset under different labeled-data budgets. The image was preprocessed to $288 \times 288 \times 160$ and $0.5$mm isotropic resolution. A larger decoder~(58M trainableparameters) improves test Dice by only 0.2 (87.3 vs.\ 87.1; Supplementary Section~7.3), so the knee results do not depend on decoder capacity.

\begin{figure}[!th]
\centering
\includegraphics[width=1.0\textwidth]{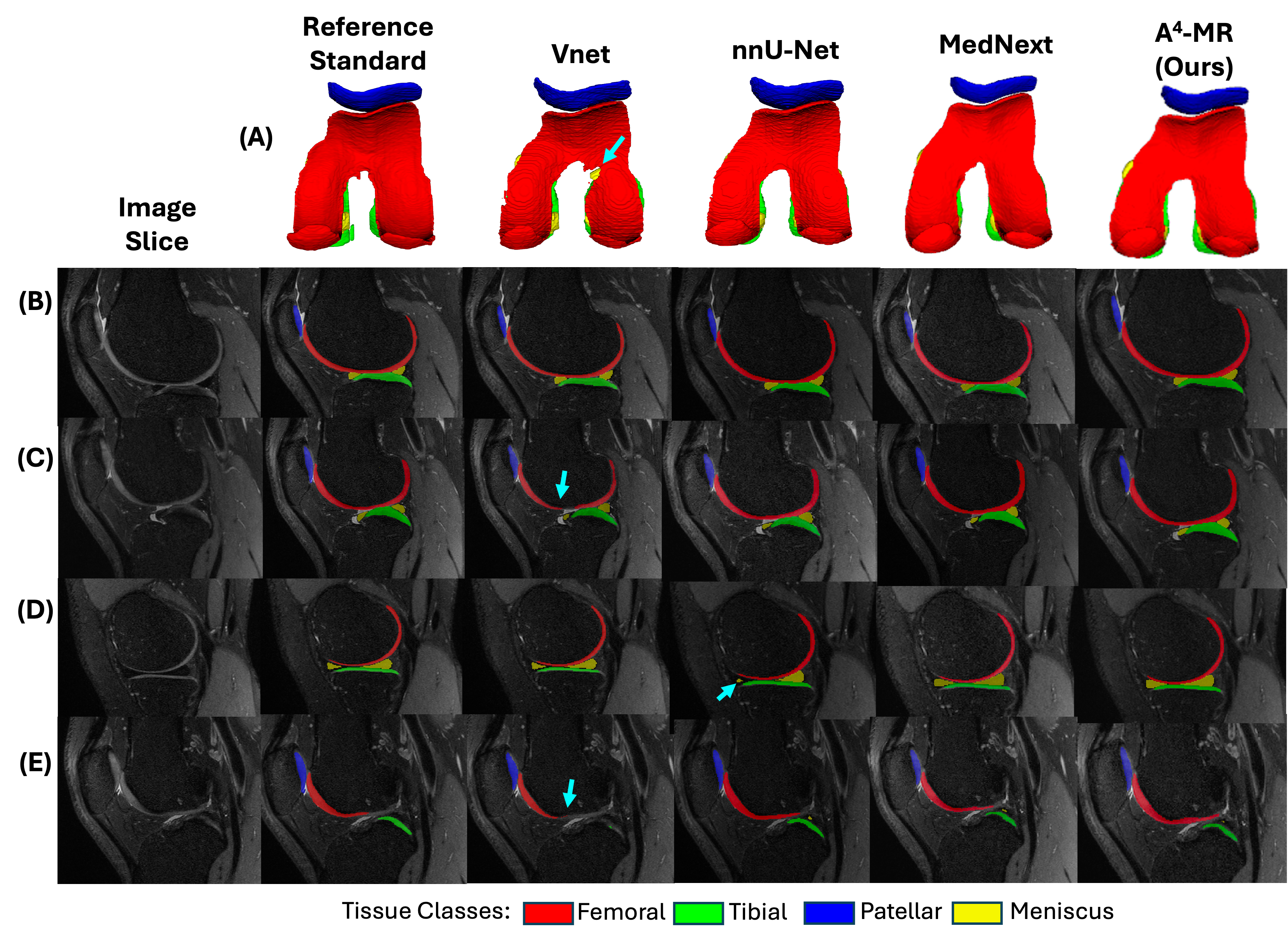}
\caption{Qualitative assessment from an external clinical case never seen by the models during training. (A) 3D rendering of segmentation outputs from the reference standard and all models in axial views. (B), (C),(D)and (E) show different sagittal slices from the same case with the respective overlaid segmentations for femoral, patellar and tibial cartilages and the meniscus horns. In (B) and (C), strong agreement is observed on sagittal slices through the center of the structures. However, in peripheral slices (D) and (E), models begin to disagree, particularly for femoral cartilage and the meniscus. Cyan arrows show bigger missegmentation errors.}
\label{fig:knee-demo}
\end{figure}

\begin{table}[!h]
    \centering
    \caption{Knee segmentation performance (Dice in \%, higher is better). {Bold: best; underlined: second best.}}
    \label{tab:knee_seg}
    \setlength{\tabcolsep}{4.5pt}
    \begin{tabular}{lccccc}
        \toprule
        Model & Femoral Cartilage & Tibial Cartilage & Patellar Cartilage & Meniscus & Average \\
        \midrule
        A\textsuperscript{4}-MR (ours) & \textbf{87.5} & \textbf{86.6} & \textbf{88.7} & \textbf{85.6} & \textbf{87.1} \\
        MedNeXt & \underline{86.6} & \underline{85.0} & 87.4 & \underline{84.4} & \underline{85.9} \\
        VNet & 86.4 & \underline{85.0} & \underline{88.0} & 84.1 & \underline{85.9} \\
        nnUNet3D & 79.9 & 79.2 & 86.8 & 79.7 & 81.4 \\
        \bottomrule
    \end{tabular}
\end{table}

\begin{figure}[!th]
\centering
\includegraphics[width=0.8\textwidth]{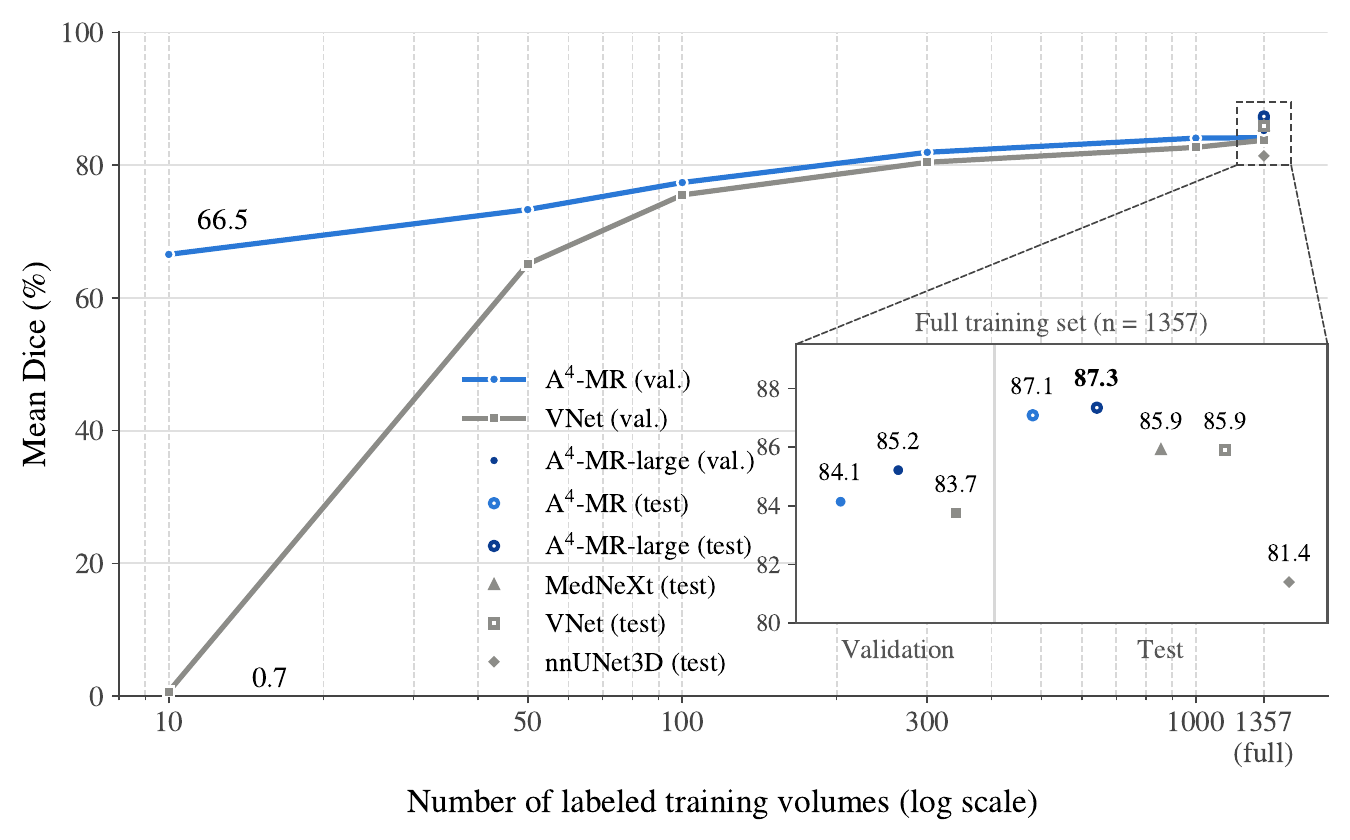}
\caption{Label efficiency on knee segmentation: mean Dice over femoral, tibial and patellar cartilage and the meniscus versus the number of labeled training volumes (log scale). A\textsuperscript{4}-MR uses a frozen encoder and the lightweight decoder; VNet is trained end-to-end. Curves (filled markers) are on the validation set. At the full training set, the inset compares these validation end points with test-set results from Table~\ref{tab:knee_seg} (hollow markers), including MedNeXt and nnUNet3D, which were trained on the full set only, and A\textsuperscript{4}-MR with a larger 58M decoder (A\textsuperscript{4}-MR-large; Supplementary Section~7.3).}
\label{fig:knee-data-suff}
\end{figure}

Figure~\ref{fig:knee-demo} shows a qualitative assessment of knee cartilage segmentation and Table~\ref{tab:knee_seg} summarizes segmentation accuracy {for femoral, tibial and patellar cartilage and the meniscus on the test set. A\textsuperscript{4}-MR achieves Dice scores of 87.5, 86.6, 88.7 and 85.6, respectively, yielding an average Dice of 87.1. Compared with fully supervised end-to-end baselines, A\textsuperscript{4}-MR is best on every structure and on average: MedNeXt~\cite{Roy2023MedNeXtTS} and VNet both obtain an average Dice of 85.9, and nnUNet3D 81.4.}

Figure~\ref{fig:knee-data-suff} shows validation Dice for $N \in \{10, 50, 100, 300, 1000\}$ labeled volumes and the full set of 1,357 volumes. A\textsuperscript{4}-MR consistently outperforms the fully supervised VNet baseline, particularly in the low-data regime. With only $N=10$ labeled cases, A\textsuperscript{4}-MR
reaches an average Dice of 66.5, whereas VNet is near zero ({0.7}). With $N=50$ and $N=100$, A\textsuperscript{4}-MR reaches 73.3 and 77.4 Dice, exceeding VNet ({65.1 and 75.5}). The gap narrows as labels increase, as expected for an end-to-end model trained on many labels (84.1 vs.\ 83.7 on the validation set with all 1,357 volumes; overlapping 95\% confidence intervals), and on the test set A\textsuperscript{4}-MR remains ahead (87.1 vs.\ 85.9).

Overall, these results indicate that A\textsuperscript{4}-MR supports accurate and label-efficient segmentation on internal knee cartilage data, where the model consistently improves over fully supervised baselines, and substantially reduces the amount of labeled data required to reach competitive performance.

With a frozen encoder and a lightweight decoder, A\textsuperscript{4}-MR matches or exceeds strong task-specific models across anatomies. It is within 0.4 of the best model on AMOS (85.3 vs.\ 85.7),
is best on CirrMRI600+ (96.5), and stays competitive on Dataset A (87.6 vs.\ 88.7). On knee segmentation, it beats end-to-end MedNeXt and VNet (87.1 vs.\ 85.9 average Dice for both).
Its largest advantage is in the low-label regime: with only 10 annotated cases it reaches $66.5$ Dice, while VNet fails to converge. Together, these results show that A\textsuperscript{4}-MR's pretrained representations carry
fine-grained anatomical information. This lets accurate, label-efficient segmentation be achieved without fine-tuning the encoder.

\FloatBarrier
\subsection{Localization and Detection}
\label{sec:localization}

We evaluate 3D organ localization on the AMOS22 MR validation set (247 image--organ pairs after exclusions; see Datasets in Section~\ref{sec:results}). A\textsuperscript{4}-MR predicts a 3D bounding box for each organ directly from the full volume, without prompts or task-specific fine-tuning; we report the same for its A\textsuperscript{4}-MR-CLIP variant. Localization is measured by the 3D intersection-over-union (IoU) between predicted and ground-truth boxes, averaged over images per organ and then over the 13 organs. Preprocessing and metric details are given in Supplementary Section~7.4.

To our knowledge, no public model predicts organ bounding boxes directly from full 3D MRI volumes. We therefore compare against MedSAM2~\cite{medsam2}, a promptable medical foundation model that propagates 2D box prompts through the volume. It receives ground-truth boxes on one slice (1 box) or on three slices (3 slices) per organ, so its results show what prompted segmentation can achieve rather than fully automatic localization. As an upper bound, we use the nnUNet3D trained on AMOS from Section~\ref{sec:segmentation} and convert its predicted masks to bounding boxes.

Table~\ref{tab:bbox_iou_exclusion} summarizes the results. Without any prompt, A\textsuperscript{4}-MR (66.3) slightly outperforms MedSAM2 given a ground-truth box (65.5), while MedSAM2 with three prompted slices (74.8) and the AMOS-trained nnUNet3D (82.4) remain stronger. On AMOS, the abnormality-aware loss also improves localization (66.3 vs.\ 65.0 without it), with gains on 11 of the 13 organs and up to +4.4 IoU on the left adrenal gland; the only decrease, on the inferior vena cava, is 0.5. 
For large, compact organs (liver, spleen, kidneys), a single prompted box can cover most of the extent, and MedSAM2 reaches 90--97 IoU versus 75--81 for A\textsuperscript{4}-MR. For elongated or hollow organs that span many slices with changing cross-sections, slice-wise propagation struggles: A\textsuperscript{4}-MR outperforms MedSAM2 with one box on the esophagus (61.2 vs.\ 38.4), stomach (76.0 vs.\ 51.4), inferior vena cava (70.7 vs.\ 47.9) and duodenum (60.9 vs.\ 41.1), and even MedSAM2 with three slices on the first three. Small organs remain hardest for all models; for the adrenal glands and the gallbladder, A\textsuperscript{4}-MR reaches only 42--52 IoU, consistent with the 8-voxel patch size of its encoder. Figure~\ref{fig:bbox_examples} shows qualitative examples for volumes with the best, median and worst localization performance.

\begin{table}[!ht]
\centering
\caption{Mean 3D bounding-box IoU (\%) on AMOS22 MR validation after excluding unreliable
(image, organ) pairs ($n{=}247$). Scores are averaged over images per organ, then over the
13 organs. MedSAM2 is prompted with ground-truth boxes on one slice (1 box) or on three
slices per organ (3 slices). nnUNet3D is trained on AMOS; all other models are
evaluated zero-shot. Bold: best; underlined: second best
among the models above the rule; the AMOS-trained reference is not ranked.}
\label{tab:bbox_iou_exclusion}
\begin{tabular}{lc}
\toprule
Model & Bounding-box IoU \\
\midrule
A\textsuperscript{4}-MR (zero-shot)  & \underline{66.3} \\
A\textsuperscript{4}-MR-CLIP (zero-shot)       & 65.0 \\
\midrule
MedSAM2 (1 box)      & 65.5 \\
MedSAM2 (3 slices)   & \textbf{74.8} \\
\midrule
nnUNet3D (AMOS-trained) & 82.4 \\
\bottomrule
\end{tabular}
\end{table}

\begin{figure*}[!htbp]
\centering
{\offinterlineskip
\makebox[1.1em]{}\hspace{2pt}%
\makebox[0.31\textwidth][c]{\small Best (amos\_0556)}\hspace{0pt}%
\makebox[0.31\textwidth][c]{\small Median (amos\_0561)}\hspace{0pt}%
\makebox[0.31\textwidth][c]{\small Worst (amos\_0553)}\\[3pt]
\parbox[b][0.2518\textwidth][c]{1.1em}{\centering\rotatebox{90}{\small Liver}}\hspace{2pt}%
\includegraphics[width=0.31\textwidth]{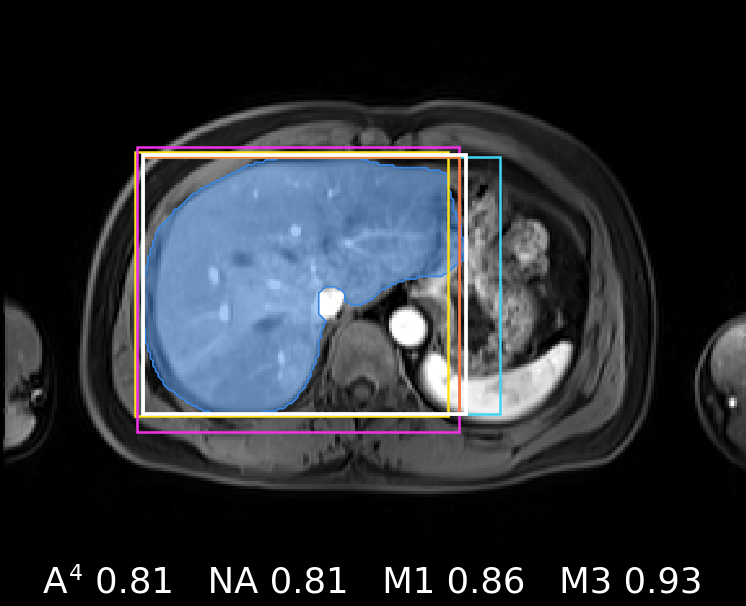}\hspace{0pt}%
\includegraphics[width=0.31\textwidth]{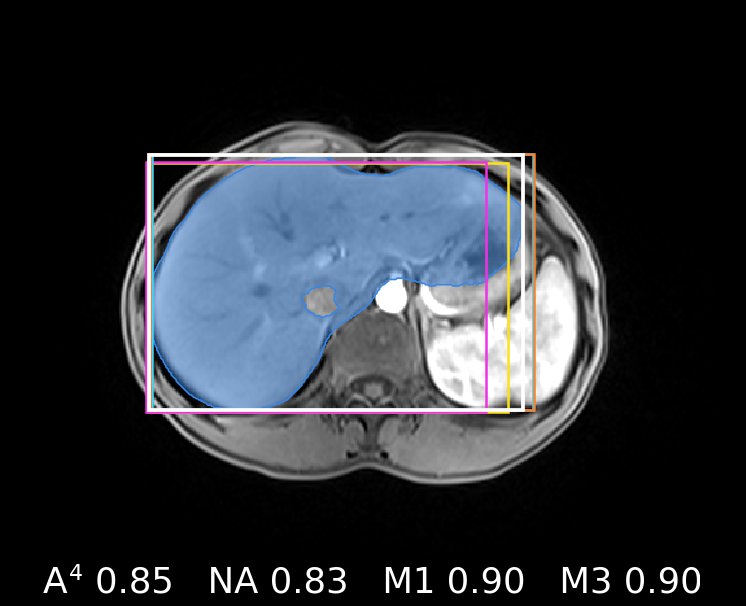}\hspace{0pt}%
\includegraphics[width=0.31\textwidth]{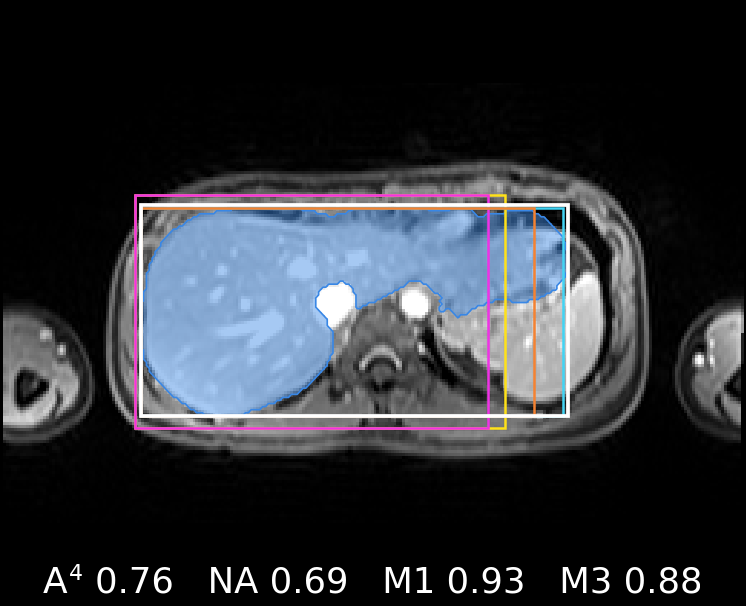}\\[0pt]
\parbox[b][0.2518\textwidth][c]{1.1em}{\centering\rotatebox{90}{\small Spleen}}\hspace{2pt}%
\includegraphics[width=0.31\textwidth]{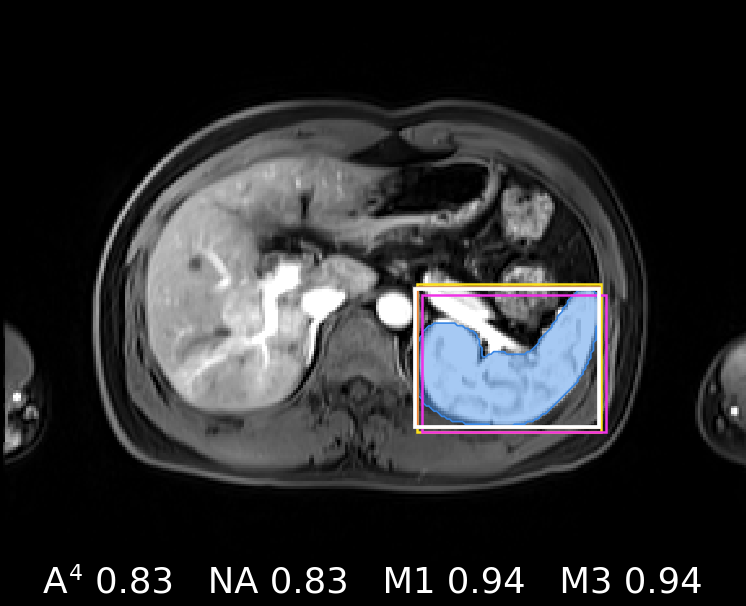}\hspace{0pt}%
\includegraphics[width=0.31\textwidth]{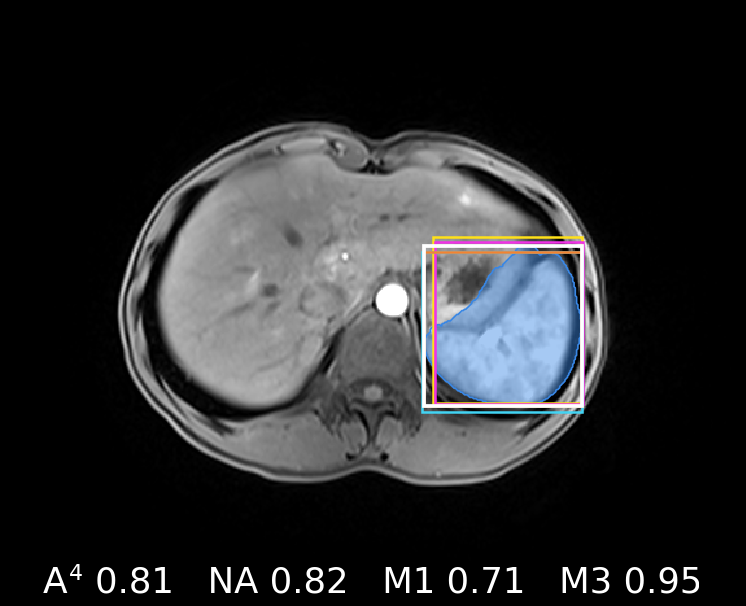}\hspace{0pt}%
\includegraphics[width=0.31\textwidth]{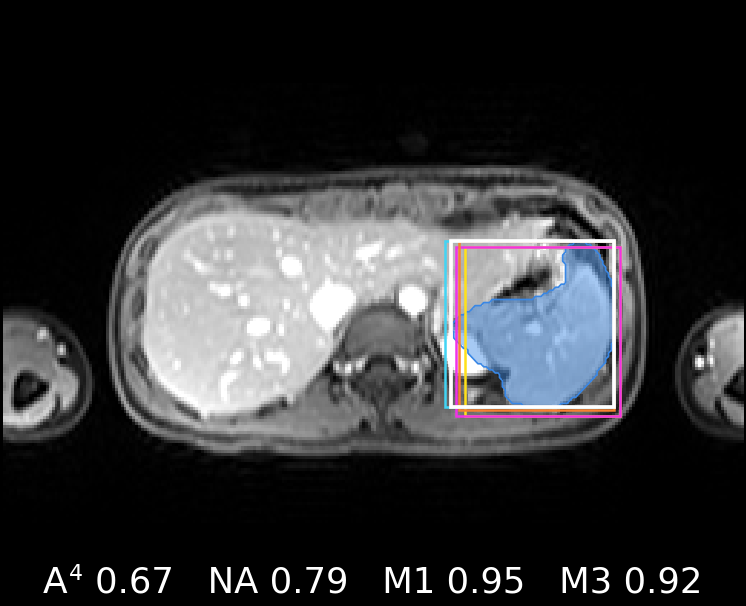}\\[0pt]
\parbox[b][0.2518\textwidth][c]{1.1em}{\centering\rotatebox{90}{\small Pancreas}}\hspace{2pt}%
\includegraphics[width=0.31\textwidth]{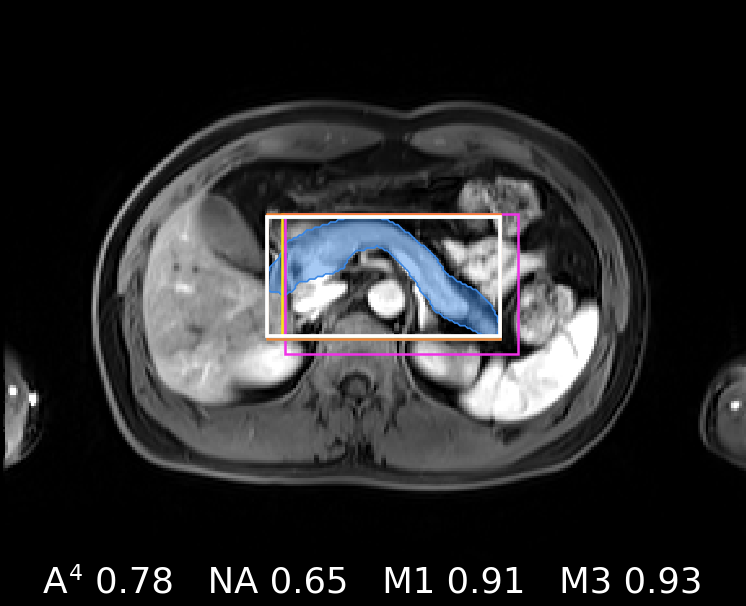}\hspace{0pt}%
\includegraphics[width=0.31\textwidth]{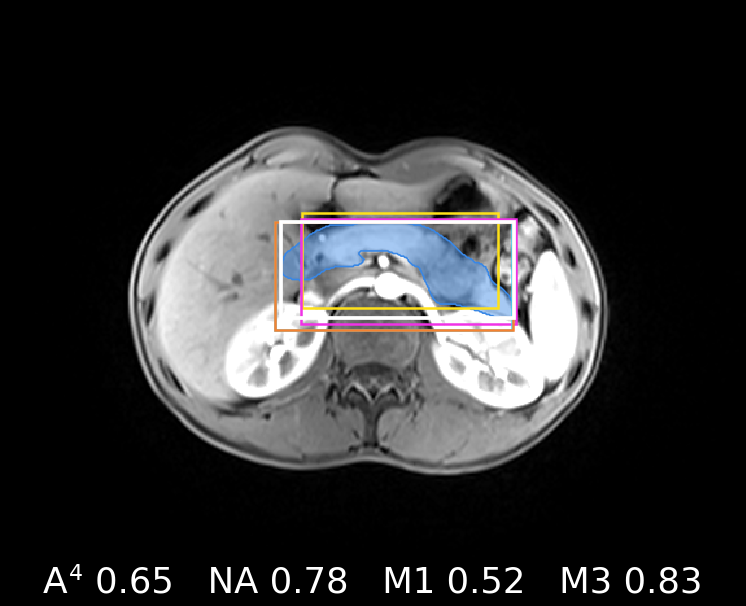}\hspace{0pt}%
\includegraphics[width=0.31\textwidth]{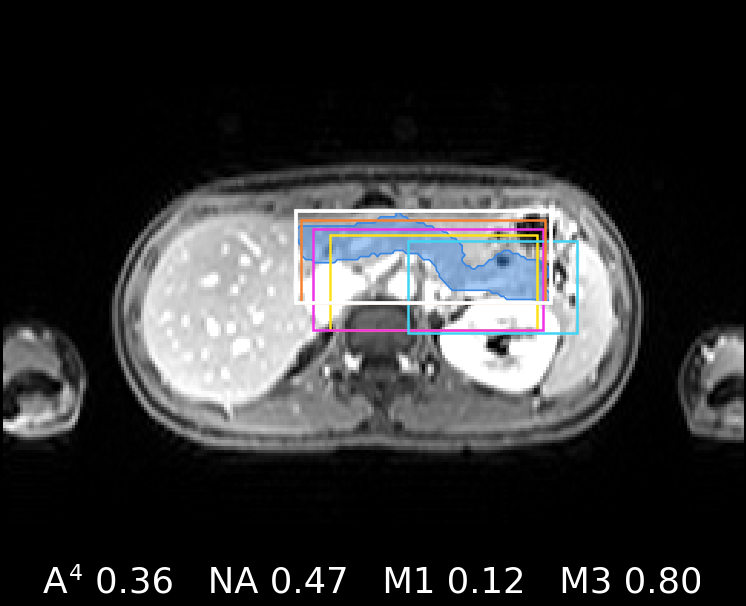}\\[0pt]
\parbox[b][0.2518\textwidth][c]{1.1em}{\centering\rotatebox{90}{\small Stomach}}\hspace{2pt}%
\includegraphics[width=0.31\textwidth]{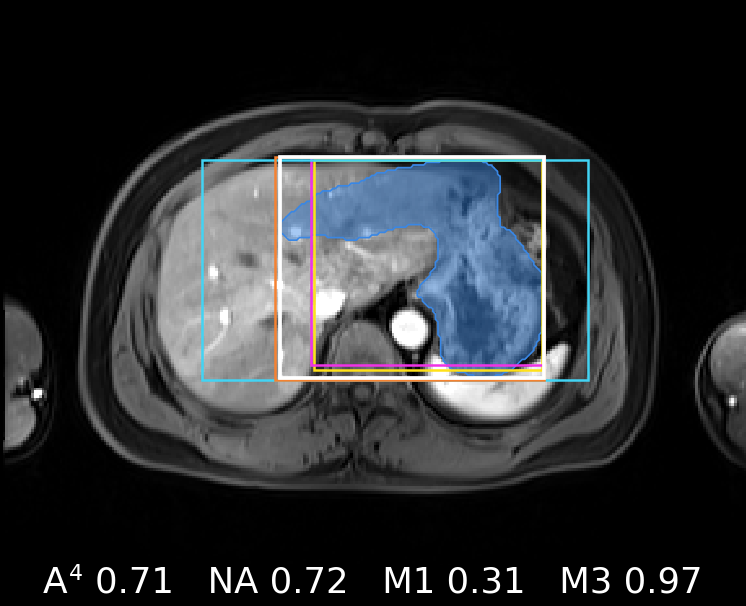}\hspace{0pt}%
\includegraphics[width=0.31\textwidth]{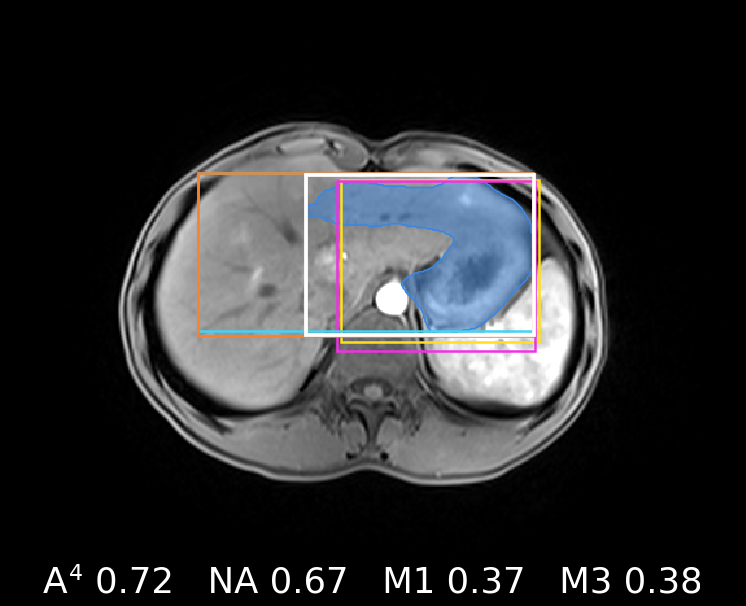}\hspace{0pt}%
\includegraphics[width=0.31\textwidth]{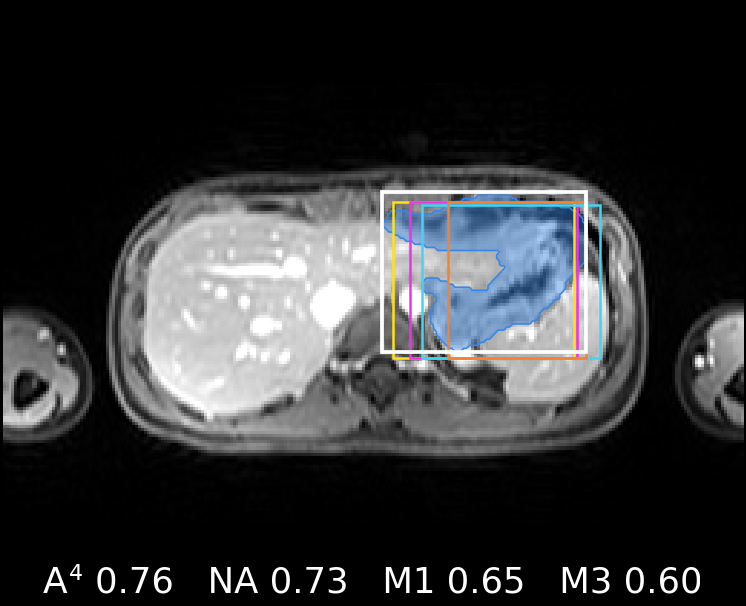}\\[0pt]
\parbox[b][0.2518\textwidth][c]{1.1em}{\centering\rotatebox{90}{\small Right adrenal}}\hspace{2pt}%
\includegraphics[width=0.31\textwidth]{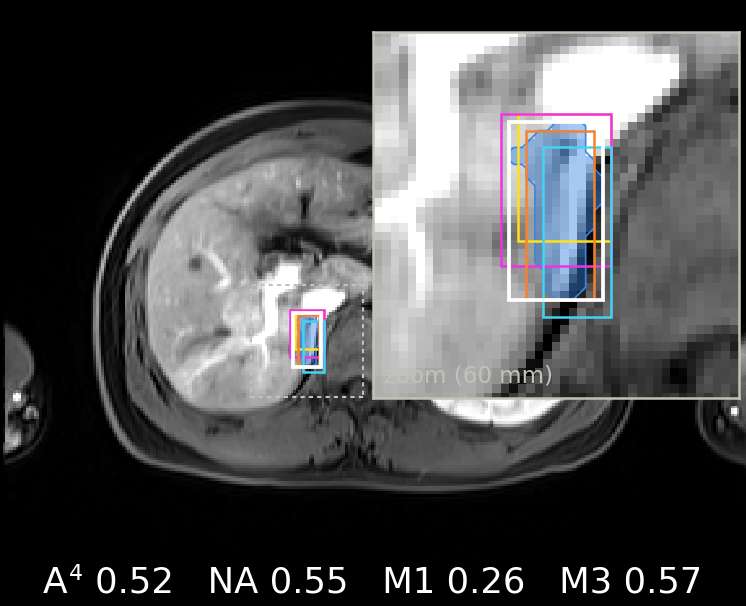}\hspace{0pt}%
\includegraphics[width=0.31\textwidth]{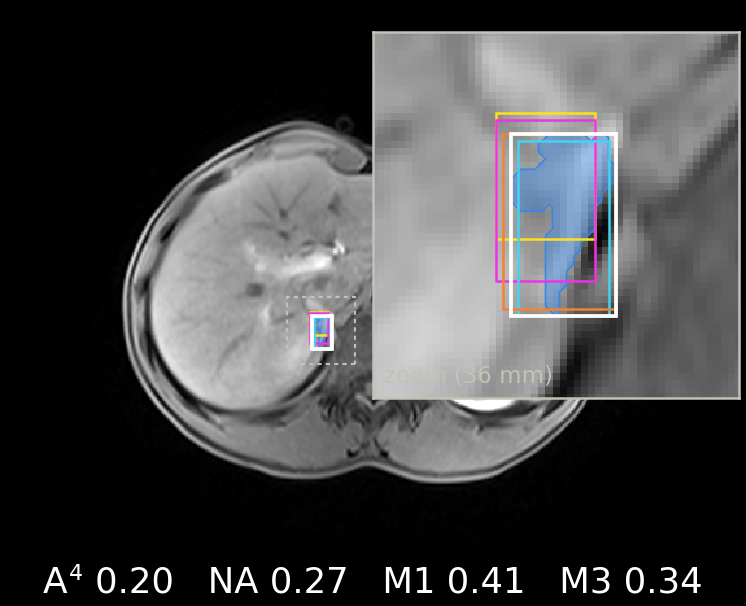}\hspace{0pt}%
\includegraphics[width=0.31\textwidth]{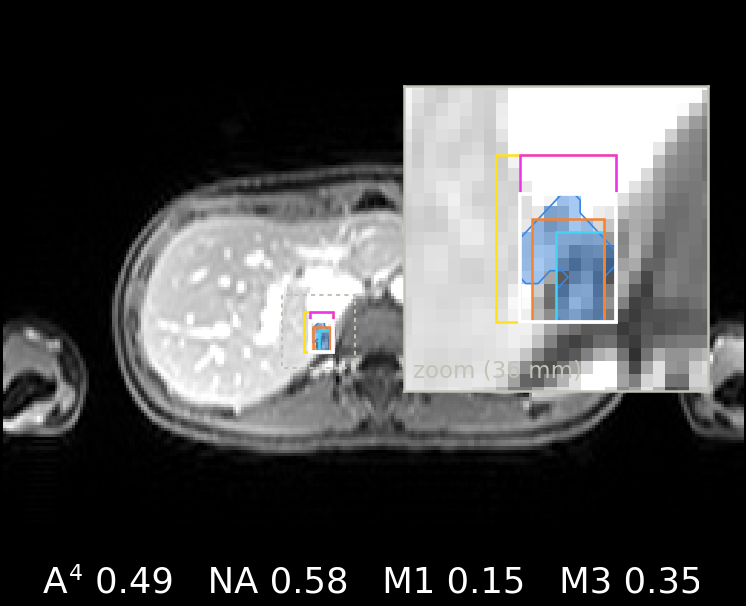}\\[6pt]
}
\includegraphics[width=0.8\textwidth]{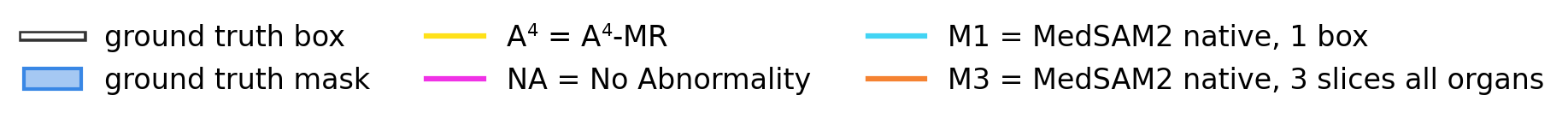}
\caption{Ground-truth and predicted 3D bounding boxes on AMOS22 MR validation for the
volumes with the best, median and worst mean bounding-box IoU, averaged over all models and
organs (columns), for five organs (rows). Each panel is the axial slice on which the
ground-truth mask best fills its bounding box. The ground-truth mask (blue), the
ground-truth box (white) and each model's box (colored) are projected along the head--foot
axis onto this slice. Numbers at the bottom of each panel give the bounding-box IoU per
model (A\textsuperscript{4}: A\textsuperscript{4}-MR; NA: A\textsuperscript{4}-MR-CLIP; M1, M3: MedSAM2 with one or three box prompts). The
right adrenal gland is shown enlarged in an inset. Volumes whose shown organs contain stray
label voxels (Supplementary Fig.~10) were not considered when selecting
these examples.}
\label{fig:bbox_examples}
\end{figure*}

\FloatBarrier
{\subsection{Prostate Report Generation}
\label{sec:report_gen}

The frozen encoder can also condition a language decoder. We connect it to a frozen LLM and generate structured prostate
MR reports. We ask for a structured JSON report rather than free text so that each finding can be scored directly
against the radiologists' labels; free-text overlap metrics reward wording more than clinical correctness. Following a
prefix-based setup~\cite{mokady2021clipcap}, the encoder embeds each series (T2W, ADC, DWI), the three embeddings are
averaged, and a single trained linear layer with LayerNorm maps them into the embedding space of a frozen
Qwen2.5-3B-Instruct decoder~\cite{qwen2025qwen25technicalreport}, where they are prepended to the prompt. The projection
is trained on 250 Dataset B studies and the checkpoint is selected on 50 other studies; we report mean $\pm$ SD over
the three of five training seeds with the best validation score (all seeds in Supplementary Tab.~17). We compare with Decipher-MR in the same setup and with zero-shot vision-language models given the same
prompt (Qwen2.5-VL~\cite{Bai2025Qwen25VLTR}, 3B and 32B, and MedGemma-1.5-4B~\cite{Sellergren2025MedGemmaTR}). On a
balanced held-out test set of 60 studies (30 PI-RADS $\le$2 and 10 each of PI-RADS 3, 4 and 5), we score two study-level
findings: PI-RADS $\ge$4, i.e.\ a lesion for which clinically significant cancer is likely~\cite{Weinreb2015PIRADSPI},
and the prostate volume (details in Supplementary Sec.~7.5).

With the trained projection, A\textsuperscript{4}-MR gives a weak signal for PI-RADS $\ge$4 (balanced accuracy $60.0 \pm 5.7$, vs.\
$56.7 \pm 3.1$ for Decipher-MR; zero-shot models at or below chance, 50) and reports the prostate volume with an error
of 16.7\,cc (Decipher-MR 17.9\,cc), below the 21.8\,cc of always answering the training median (Tab.~\ref{tab:report_test60}). The zero-shot
models do not give usable volumes: Qwen2.5-VL-3B answers 1{,}000 or 1{,}800\,cc, and the other models rarely give a
volume. The prostate box detected by A\textsuperscript{4}-MR gives a better volume still, without any language model (ellipsoid
volume, 11.1\,cc). An image prefix averaged over the series thus carries organ size but little lesion information;
conditioning the decoder on localized tokens, such as the box pooling of Sec.~\ref{sec:results}, is a natural next step.
\par}

\begin{table}[!htbp]
\centering
\small
\setlength{\tabcolsep}{4pt}
\caption{Prostate report generation on a balanced held-out test set (60 Dataset B studies: 30 PI-RADS $\le$2 and
10 each of PI-RADS 3, 4 and 5). PI-RADS $\ge$4: balanced accuracy (BA) against the radiologists' study-level score
(any constant answer scores 50). Prostate volume: number of studies with a volume answer and mean absolute error (cc)
against the volume stated in the report. Trained rows: frozen encoder, trained linear projection, frozen Qwen2.5-3B;
mean $\pm$ SD over the three of five training seeds with the best validation score, the same for both encoders (seeds 42, 1, 2); all seeds in Supplementary Tab.~17.}
\label{tab:report_test60}
\begin{tabular}{lccc}
\toprule
 & PI-RADS $\ge$4 & \multicolumn{2}{c}{Prostate volume} \\
\cmidrule(lr){3-4}
Method & BA & answered & MAE (cc) \\
\midrule
\multicolumn{4}{l}{\textit{Zero-shot vision-language models}} \\
Qwen2.5-VL-3B          & 50.0 & 52 & 1233 \\
Qwen2.5-VL-32B         & 48.8 & 1  & 43.0 \\
MedGemma-1.5-4B (3D)   & 43.8 & 22 & 35.2 \\
\midrule
\multicolumn{4}{l}{\textit{MR encoder + Qwen2.5-3B (trained projection)}} \\
Decipher-MR            & 56.7$\pm$3.1 & 59.3$\pm$0.6 & 17.9$\pm$1.1 \\
A\textsuperscript{4}-MR                    & 60.0$\pm$5.7 & 60.0$\pm$0.0 & 16.7$\pm$4.2 \\
\midrule
\multicolumn{4}{l}{\textit{References}} \\
Constant answer (training median) & 50.0 & 60 & 21.8 \\
A\textsuperscript{4}-MR box ellipsoid (no LLM) & -- & 60 & 11.1 \\
\bottomrule
\end{tabular}
\end{table}

\subsection{Subgroup Analysis}
\label{sec:fairness}
\begingroup
We test whether frozen-encoder probes perform equally across patient and acquisition subgroups on ADNI (AD vs CN,
MCI vs CN, brain age), Dataset B (prostate) (clinically significant cancer, ISUP$\geq$2) and CirrMRI600+ (cirrhosis vs
healthy). Attributes are sex, age (younger $<$70 vs older $\geq$70 years in every dataset), scanner vendor and race
(Dataset B), and field strength on ADNI (1.5\,T in ADNI1, 3\,T in ADNI2/GO; the phases also differ in processing). Probes follow each
dataset's protocol (ADNI: 50 random splits; Dataset B: stratified 5-fold cross-validation repeated 5 times; CirrMRI600+:
patient-grouped 5-fold cross-validation over the full cohort, since the official test split contains only five
healthy patients), and each subject's score is averaged over the runs in which it was tested. We run two analyses:
(i) the probe trained on all subjects, evaluated per subgroup --- AUROC (MAE for brain age) and subgroup gaps, plus
sensitivity and false-positive rate at one operating point per model; (ii) probes trained on one subgroup only
(women or men; one vendor; size-matched age groups) and tested on every subgroup. All differences come with 95\%
patient-bootstrap confidence intervals and paired significance tests. Volumes are fed in the pretraining
orientation of their acquisition plane; external 2D models are applied slice-wise (24 axial slices, mean-pooled).
Details and full results are in Supplementary Sec.~8.

\begin{figure}[t]
\centering
\includegraphics[width=\textwidth]{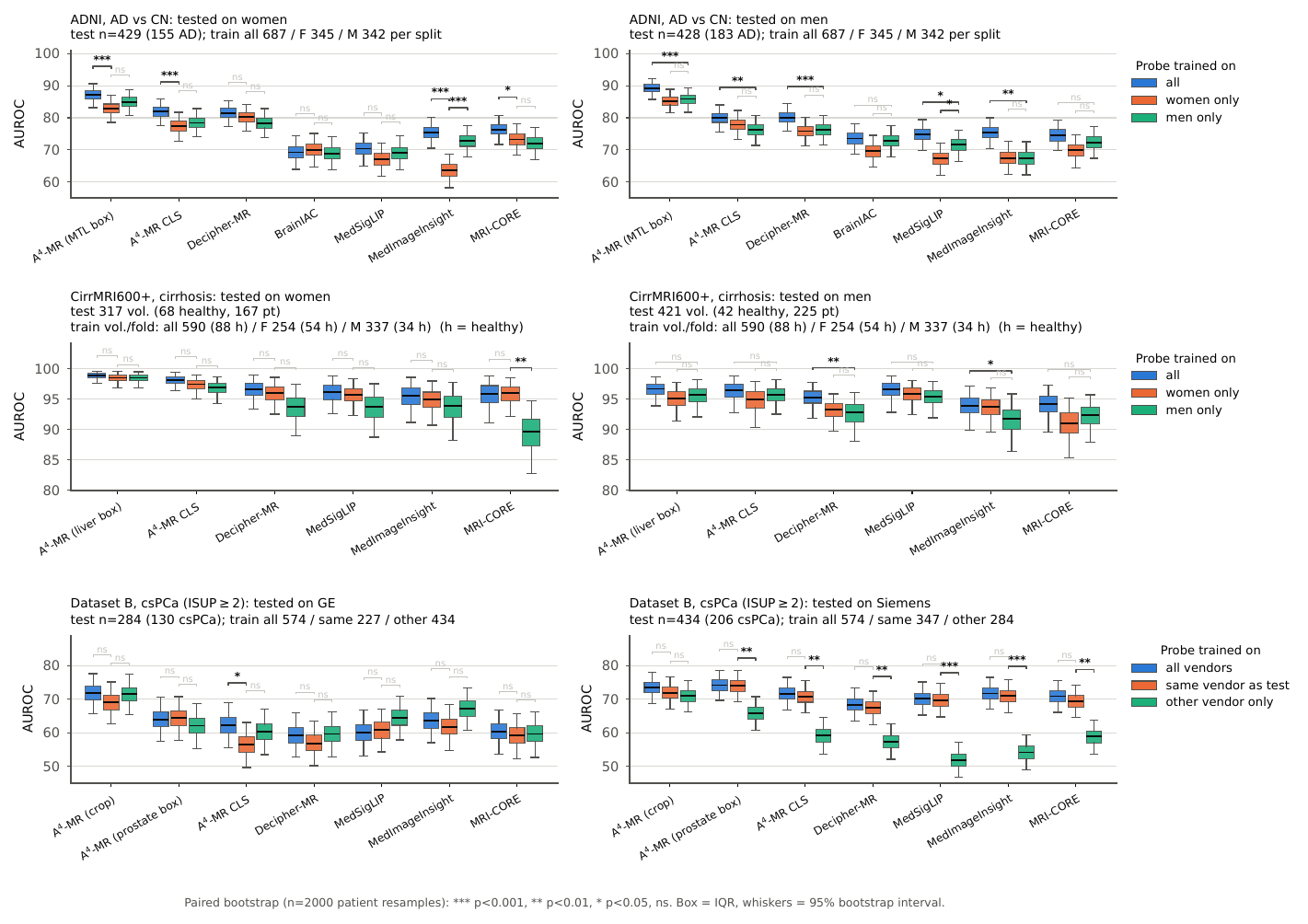}
\caption{Subgroup performance (AUROC). Each panel is one test subgroup; for each model, probes trained on
all subjects, on women or men only (ADNI, CirrMRI600+), or on all vendors, the test vendor only, or the other vendor
only (Dataset B). Boxes: interquartile range of 2{,}000 patient-bootstrap resamples, whiskers: 95\% interval. Brackets
compare training sets on the same test patients (paired bootstrap; *** $p<0.001$, ** $p<0.01$, * $p<0.05$, ns).
Panel titles give test and training sizes. Gaps between test subgroups for every attribute are summarised in
Supplementary Fig.~11.}
\label{fig:fairness}
\end{figure}

\paragraph{No detectable demographic gap.} With the probe trained on all subjects, none of the differences between
test subgroups is significant for A\textsuperscript{4}-MR on ADNI --- by sex, age group or field strength, for AD, MCI or brain age (AD
with the MTL-box representation: women 87.0, men 89.2; $<$70 / 70--79 / $\geq$80 years: 88.9 / 88.8 / 89.9) --- and
A\textsuperscript{4}-MR is the most accurate model in every subgroup (Fig.~\ref{fig:fairness}). Among the other models, only
BrainIAC (MCI, men $-$ women $+8.3$ points) and MRI-CORE (AD across age groups, 10.5 points) show significant gaps.
On CirrMRI600+, five of six models are 1--2 points lower for men (none significant; the cohort has 21 healthy men); on Dataset B, differences by race
are not detectable with 43 Black patients (Supplementary Fig.~11).

\paragraph{Training on one sex.} For A\textsuperscript{4}-MR, a probe trained on women only and one trained on men only perform the
same on either sex (AD, tested on women: 82.8 vs 84.8; on men: 85.3 vs 85.9; both n.s.), and both are below the probe
trained on everyone (87.0 and 89.2): the representation needs more data, not sex-specific training. This also holds
on CirrMRI600+ (liver box, n.s.). Two other models depend on the training sex: on ADNI, MedImageInsight's women-only
probe is 9.2 points worse on women than its men-only probe, and on CirrMRI600+ MRI-CORE's men-only probe is 6.4 points
worse on women (both significant). For brain age, a probe trained on the other sex is worse for every model
(Supplementary Fig.~12).

\paragraph{A shared threshold is not age- or protocol-neutral.} At one operating point per model, every ADNI model
flags older subjects more often --- higher sensitivity for AD but more false positives among older controls
(A\textsuperscript{4}-MR MTL box: sensitivity 57\% $\to$ 83\%, false-positive rate 0\% $\to$ 17\% from younger to older subjects)
--- and 3\,T (ADNI2/GO) subjects less often than 1.5\,T (ADNI1) subjects. 
A\textsuperscript{4}-MR has among the smallest shifts (field strength: $-14.9$ / $-10.5$ points in sensitivity / false-positive rate,
against $-15$ to $-22$ / $-19$ to $-26$ for the other models) and no shift by sex, whereas BrainIAC's sensitivity is
29.5 points higher for men (Supplementary Tab.~27). Deployment on brain MRI therefore needs
age- and protocol-aware calibration, for every model.

\paragraph{Scanner vendor is the one consistent source of disparity, and localization removes it.} On Dataset B,
whole-image representations of every model --- A\textsuperscript{4}-MR, Decipher-MR, MedSigLIP, MedImageInsight and MRI-CORE --- are
9--10 AUROC points worse on GE than on Siemens, and a probe trained on GE studies alone loses 10--18 points on Siemens
studies (Fig.~\ref{fig:fairness}, bottom). This is a domain shift rather than a data-size effect: a probe trained on
mixed-vendor data of the same size as the Siemens-only probe performs as well as it. Re-embedding the
A\textsuperscript{4}-MR-detected prostate crop removes the vendor dependence: the GE--Siemens gap falls to 1.8 points and the
cross-vendor loss to 1.0 point, neither significant, and the crop shows no vendor or race difference at a shared
threshold. 
\par\endgroup

\label{sec:subgroup}

\section{{Limitations and Future Work}}
\label{sec:limitations}

\begin{itemize}[leftmargin=0.5cm]
    \item {Weak and noisy {pretraining} supervision.} {During pretraining,} anatomy boxes are derived from TotalSegmentator, proprietary segmentation models, and DICOM-based coordinate transfer, and anatomy health status is extracted from reports by an LLM. Only a subset of the processed reports was manually reviewed, so label noise in both localization targets and abnormality labels {shapes the learned representations; downstream tasks are trained and evaluated on their own curated labels with the encoder frozen.} {The weak masks are also reduced to bounding boxes. A segmentation head trained with a masked loss during pretraining, applied only to anatomies for which a mask is available, would let the encoder learn from voxel-level shape information instead of the box extent alone.}
    \item {Patch resolution.} With $8\times8\times 8$ patches, structures smaller than a single patch along any axis cannot be resolved, and zero-shot localization still trails prompted (multi-slice) MedSAM2 and dataset-specific supervised models, particularly on small organs such as the adrenal glands. {Smaller or anisotropic patches, or multi-scale tokenization, could improve localization and segmentation of small anatomies at a manageable computational cost.}
    \item {Coarse abnormality modeling.} Abnormality is represented as a binary healthy/abnormal state per organ, propagated to all mentions of that organ within a report. This does not capture severity, multiple co-occurring findings, or lesion-level detail, and our abnormality evaluation focuses mainly on a single organ and disease (liver cirrhosis). {Abnormality is learned implicitly through the structure of the contrastive targets. A classification head on the per-anatomy decoder embeddings, trained with an explicit abnormality loss, could provide a more direct training signal, enable organ-level abnormality prediction without a downstream probe, and serve as a first step toward severity and lesion level labels.}
    \item {{Fixed anatomy list.} The localization decoder detects a fixed list of anatomy classes. Because each detected anatomy has an embedding aligned with the text encoder, this list can already be queried with free text rather than class names: a query such as ``anatomies required for vision'' matches all related classes, including the left and right eyes, lenses and optic nerves. Howeever, anatomies outside the vocabulary list cannot be localized. Extending the list, together with the corresponding weak labels and report findings, would broaden anatomy-level alignment and abnormality awareness to more body regions and structures, towards open-vocabulary setup.}
    \item {{From anatomies to pathologies.} Existing MR foundation models such as MRI-CORE~\cite{Dong2025MRICOREAF}, OmniMRI~\cite{he2025omnimri} and Decipher-MR~\cite{Yang2025DecipherMRAV} learn image-level representations, in some cases aligned with radiology reports. A\textsuperscript{4}-MR extends this alignment to anatomical regions across the body. The natural next step is to bring lesions, tumors and other pathologies into pretraining in the same way: findings extracted from reports provide weak text supervision, and existing lesion detection and segmentation models can bootstrap their locations. Any pretrained model can be fine-tuned for lesion detection or delineation; the goal here is to make pathology grounding part of the learned representation itself.}
\end{itemize}

\begin{ack}
We thank Eszter Katalin Kiss for reviewing the liver abnormality images, and the GE HealthCare Real World Data (RWD) team for their support.
\end{ack}

\printbibliography

\end{document}